\documentclass{article} %
\usepackage[preprint]{neurips_2021}

\usepackage{amsmath,amsfonts,bm}

\usepackage[utf8]{inputenc} %
\usepackage[T1]{fontenc}    %
\usepackage{hyperref}       %
\usepackage{url}            %
\usepackage{booktabs}       %
\usepackage{nicefrac}       %
\usepackage{microtype}      %
\usepackage{xcolor}         %
\usepackage{wrapfig}
\usepackage{floatrow}

\definecolor{myred}{rgb}{0.658, 0.188, 0.000}

\usepackage{booktabs}
\usepackage{graphicx}
\usepackage{epsfig}
\usepackage{multirow}
\usepackage[ruled,vlined]{algorithm2e}
\usepackage{amsmath}

\newcommand\rev[1]{\textcolor{black}{#1}}

\newcommand{\ShortName}{HAFL\xspace}

\def\eqref#1{equation~\ref{#1}}

\def\1{\bm{1}}

\DeclareMathAlphabet{\mathsfit}{\encodingdefault}{\sfdefault}{m}{sl}
\SetMathAlphabet{\mathsfit}{bold}{\encodingdefault}{\sfdefault}{bx}{n}

\usepackage{hyperref}
\usepackage{url}
\usepackage{colortbl}
\usepackage{pifont}
\usepackage{float}
\newcommand{\cmark}{\ding{51}}%
\newcommand{\xmark}{\ding{55}}%

\title{Federated Learning with Heterogeneous Architectures using Graph HyperNetworks}

\author{Or Litany, Haggai Maron, David Acuna, Jan Kautz, Gal Chechik, Sanja Fidler\\
NVIDIA\\
\texttt{\{olitany,hmaron,dacunamarrer,jkautz,gchechik,sfidler\}@nvidia.com}}

\begin{document}

\maketitle

\begin{abstract}
Standard Federated Learning (FL) techniques are limited to clients with identical network architectures. This restricts potential use-cases like cross-platform training or inter-organizational collaboration when both data privacy and architectural proprietary are required. We propose a new FL framework that accommodates heterogeneous client architecture by adopting a graph hypernetwork for parameter sharing. A property of the graph hyper network is that it can adapt to various computational graphs, thereby allowing meaningful parameter sharing across models. Unlike existing solutions, our framework does not limit the clients to share the same architecture type, makes no use of external data and does not require clients to disclose their model architecture. Compared with distillation-based and non-graph hypernetwork baselines, our method performs notably better on standard benchmarks. We additionally show encouraging generalization performance to unseen architectures. %
\end{abstract}

\section{Introduction}
Federated learning (FL) \citep{mcmahan2017communication,yang2019federated, konecny2015federated,konecny2017federated} allows multiple clients to collaboratively train a strong model that benefits from their individual data, without having to share that data. In particular, aggregating the parameters of locally trained models alleviates the need to share raw data, thereby preserving privacy to some extent and reducing the volume of data transferred. FL allows safe and efficient learning from edge nodes like smartphones and self-driving cars. Another important use case is to improve performance among medical facilities while maintaining patient privacy, for example in disease identification.

Most current FL approaches have a key limitation: all clients must share the same network architecture. As a result, they cannot be applied to many important cases that require learning with heterogeneous architectures. For example, depending on the computing power or OS version, different platforms may run different networks. For some organizations, changing models might hinder legacy expertise or pose regulatory challenges. Additionally, some organizations may wish to benefit from each other's access to data without sharing their proprietary architectures.  In all these cases, one is interested in \textit{federated learning with heterogeneous architectures} (\ShortName).

The following example illustrates why current FL approaches do not support clients with different architectures. Consider  Federated Averaging (FedAvg), perhaps the most widely used FL technique, where model parameters of different clients are averaged on a shared server \cite{mcmahan2017communication}. Unless all clients share the same structure of parameters and layers, it is not well defined how to average their weights. The problem of aggregating weights across different architecture is not specific to FedAvg, but exists with other FL approaches. 
This limitation raises the fundamental research question: \textbf{Can federated learning handle heterogeneous architectures? And can it be achieved with clients keeping their architectures undisclosed?}

\begin{figure}
\floatbox[{\capbeside\thisfloatsetup{capbesideposition={right,top},capbesidewidth=4.5cm}}]{figure}[\FBwidth]
{\caption{An overview of our approach. We tackle a federated learning setup with clients that have different architectures by using a shared Graph Hypernetwork (GHN) that can adapt to any network architecture. At each communication round, a local copy of the GHN is trained by each client using their own data and network architecture (illustrated as graphs within each client; nodes represent layers and edges represent computational flow), before returning to the server for aggregation.}\label{fig:fig1}}
    {\includegraphics[width=0.65\textwidth]{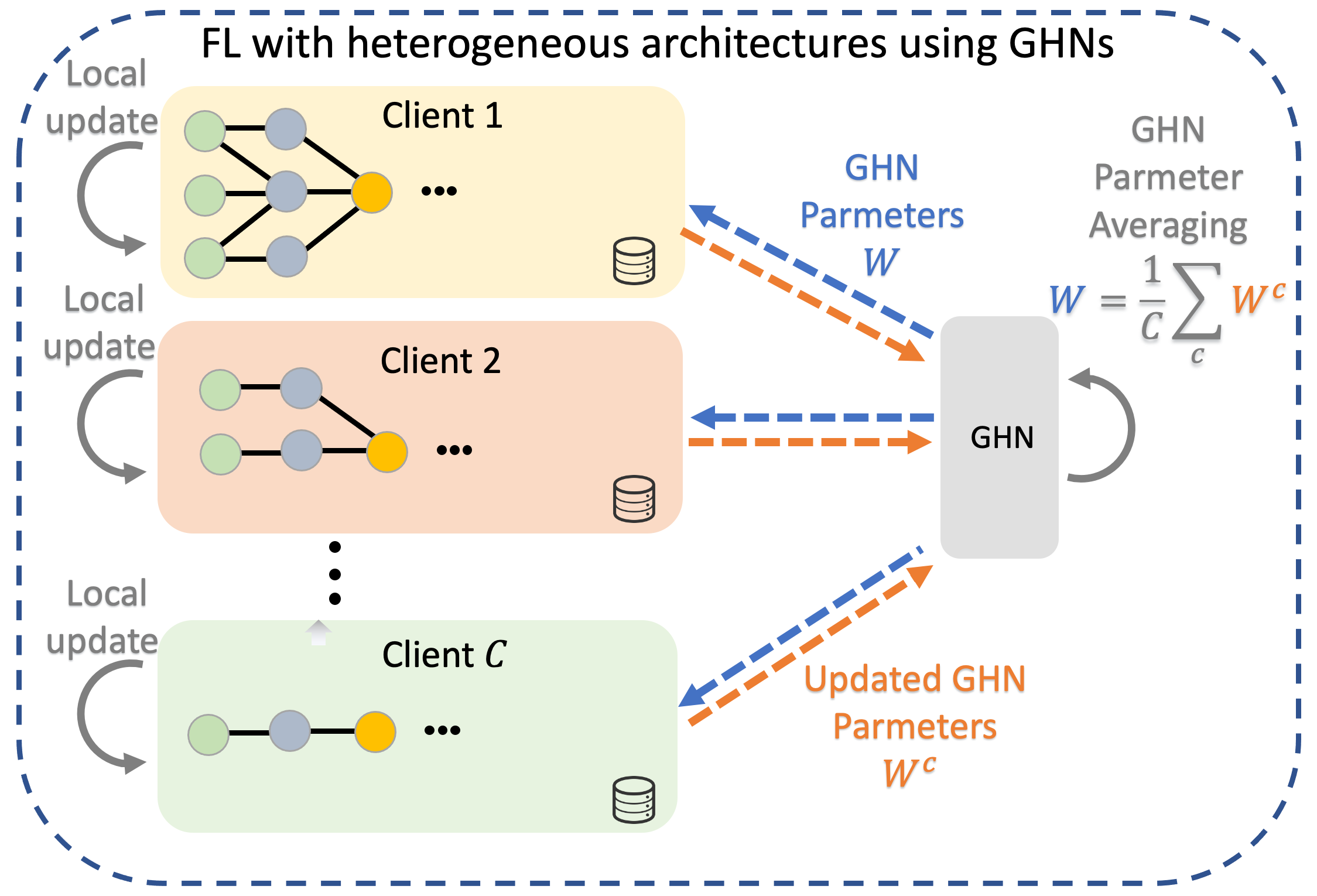}}
    \vspace{-10pt}
\end{figure}

The first work to consider FL with non-identical architectures is \cite{diao2021heterofl}. Focusing on platforms with different computation capabilities, their setup assumes all clients share the same architecture type (e.g. ResNet18) and differ only in the number of channels per layer such that smaller capacity model parameters are subsets of the larger ones. This allows averaging of corresponding parameter subsets. 
Another approach for \ShortName is knowledge distillation. For example, \cite{lin2020ensemble} suggested distilling knowledge from ensembles of client models on a server using unlabeled or synthetic datasets. Although this solution could be beneficial in certain setups, it has two major limitations: (1) Client architectures must be disclosed and (2) External data must be provided. %
We could instead design %
a distillation-based approach to \ShortName that does not require clients to share their architectures. %
First, train a global model between clients using standard FL. Then, distill knowledge from that shared model to train each client's model on local data. Even though this solution addresses the two limitations discussed above, training a large global model might be difficult for clients with limited computing power. Moreover, only using locally available data for distillation may overfit and hinder the benefits of FL. Finally, and perhaps most importantly, this approach bypasses, but does not address, the fundamental question: How can we learn to use knowledge about architectures when transferring information about model parameters between architectures?

In this work, we propose a general approach to \ShortName based on a hypernetwork (HN) \citep{ha2016hypernetworks} that acts as the knowledge aggregator. The hypernetwork takes as input a description of a client architecture and emits weights to populate that architecture. Unlike FedAvg, which aggregates weights in a predefined way, the hypernetwork implicitly learns an aggregation operator from data and can therefore be significantly more powerful.  
For providing an architecture descriptor to the HN, we propose to represent client architectures as graphs, where nodes can represent both parametric layers (such as convolutional layers) and non-parametric layers (such as a summation operation) and directional edges represent network connectivity such as skip connections. 

To allow the hypernetwork to process any graph, regardless of its size or topology, we propose to use a \textit{graph hypernetwork} (GHN) ~\citep{zhang2020graph}. We therefore name our approach \textit{\ShortName-GHN}. The GHN operates on a graph representation of the architecture and predicts layer parameters at each node that represents a parametric layer. During training, at each communication round, clients train local copies of the GHN weights using their own architectures and data (Fig. 1). Then, these weights are sent by the clients to the server where they are aggregated. Since different architectures have different layer compositions, representing layers as nodes allows meaningful knowledge aggregation across architectures. This forms an improved hypernetwork model that uses knowledge gleaned from different network types and datasets, to populate them with improved parameters. Critically, client architectures are not communicated. %

\ShortName-GHN relies on the ability of the Graph Neural Network (GNN) to generalize across different client architectures.  We discuss this matter in light of recent results in theory of GNNs~\citep{yehudai2021local} and demonstrate experimentally that our approach generalizes to unseen architectures in certain cases -- allowing clients to modify their architectures after federation has occurred, without the need for client-wide FL retraining. 

Our experimental study on three image datasets shows that \ShortName-GHN outperforms a distillation-based baseline, and a non-GNN based HN architecture~\citep{shamsian2021personalized}, by a large margin. This margin increases as the local dataset size decreases. We further show that \ShortName-GHN provides a large benefit in generalization to unseen architectures, improving the accuracy of converged models and drastically shortening the time needed for convergence. Such generalization could allow new clients, with new architectures, to benefit from models trained on different data and different architectures, lowering the bar for deploying new, personalized, architectures.

\section{Previous work}

\paragraph{Federated Learning.} 
Federated learning \citep{mcmahan2017communication, kairouz2019advances, yang2019federated,mothukuri2021survey} is a learning setup in which multiple clients collaboratively train individual models while trying to benefit from the data of all the clients without sharing their data. The most well known FL technique is federated averaging~\citep{mcmahan2017communication}, where all clients use the same architecture, which is trained locally by each client and then sent to the server and averaged with other locally trained models. Many recent works have focused on improving privacy \citep{mcmahan2017learning,agarwal2018cpsgd,li2019privacy,zhu2020federated} and communication efficiency \citep{chen2021communication,agarwal2018cpsgd,chen2021communication,dai2019hyper,stich2018local}. Another widely studied setup is the heterogeneous client data setup \citep{hanzely2020federated,zhao2018federated,sahu2018convergence,karimireddy2020scaffold,zang2021personalized}. To solve this problem, \textit{personalized FL} (pFL) methods were proposed that adapt global models to specific clients \citep{kulkarni2020survey}. FL with heterogeneous client architectures is, however, still underexplored. \cite{diao2021heterofl} considered a setup where clients share the same computational graph and differ only in the number of channels per layer. By enforcing an inclusion structure, models of different capacity can aggregate  corresponding parameter subsets. Most related to the current work is the recent work of \cite{shamsian2021personalized} that also used hypernetworks. Their method considers a simple case of clients with varying architectures, where three pre-defined simple architectures are ``hard" encoded into the structure of the hypernetwork. \rev{In contrast, our framework allows clients to use a variety of layers and computational graphs, and facilitates  better weight sharing as the same layers are used in different architectures. In the experimental section \ref{sec:exp} we show that our approach outperforms  \cite{shamsian2021personalized} by large margins.  }

\paragraph{Hypernetworks.} HyperNetworks (HN) \citep{klein2015dynamic,ha2016hypernetworks} are neural networks that predict input conditioned weights for another neural network that performs the task of interest. HNs are widely used in many learning tasks such as generation of 3D content \cite{littwin2019deep,sitzmann2019scene}, neural architecture search \cite{brock2017smash} and language modelling \cite{suarez2017character}. More relevant to our work are Graph Hypernetworks (GHNs) - hypernetworks that take graphs as an input. \cite{nachmani2020molecule} used GHNs for molecule property prediction. Even more relevant is the use of GHNs for Neural Architecture Search (NAS) \cite{zhang2020graph}. In our work, we adapt GHNs \rev{to \ShortName} with unique challenges arising from the problem setup. %

\section{Problem Definition}
Traditional FL addresses the setup of $C$ clients working together to improve their local models. Each client $c \in \{1,\dots,C \}$ has access only to its local data samples $\{(x_{cj},y_{cj})\}_{j=1}^{n_c}$, sampled from client specific data distributions $\mathcal{P}_c$, $c=1,\dots,C$.

Here, we generalize FL to FL with Heterogeneous Architectures (\ShortName). In this setup, each client can use a different network architecture $f_1(\cdot ;\theta_1), f_2(\cdot ;\theta_2), \dots, f_C(\cdot ;\theta_C)$. Here, $f_c(\cdot;\theta_c)$ \rev{is a neural architecture from some predefined family of models with learnable weights $\theta_c\in \mathbb{R}^{m_c}$ (see further discussion on architecture families in Section \ref{sec:approach}) }.  Moreover, we assume that all clients are connected to a server, and that the clients can share information among themselves only through the server. Importantly, in \ShortName we are interested in keeping both the architectures and the data private and avoid transferring it.

Our goal is to solve the following minimization problem:
\begin{equation}
    (\theta_1^*,...,\theta_C^*) = \text{argmin}_{(\theta_1,...,\theta_C)} \sum_{c=1}^C\mathbb{E}_{{(x,y) }\sim \mathcal{P}_c} \ell(f_c(x;\theta_c),y) ,
\end{equation}
for a suitable loss function $\ell(y,y')$. %

\section{Approach}\label{sec:approach}
\subsection{Overview and workflow} 
Standard FL methods rely on aggregating model parameters and are not directly applicable when clients use different architectures. In our \ShortName setup, the parameter vectors $\theta_c$ of different clients have different shapes and sizes, and as a result, a direct aggregation can be meaningless or not well defined.
To address this issue, we re-parameterize the weights $\theta_c$ of a model $c$ as the output of a hypernetwork, which serves as a trainable knowledge aggregator. The hypernetwork weights, $W$,    are learned from data using updates from all clients, and are the only weights that are trained in our model (in contrast to $\theta_c$ which are functions of $W$). Since the output of the hypernetwork must fit a given  architecture $f_c$, the hypernetwork should take as input a representation of $f_c$. 
Here, we advocate graphs as a natural representation for neural architectures (in agreement with ~\citep{zhang2020graph}) and apply graph hypernetworks to process them.  

The workflow of training our \ShortName-GHN is illustrated in Figure \ref{fig:fig1}.  At each communication round, several steps are followed. (1) First, a server shares the current weights of the GHN, $W$, with all clients (blue dashed line); (2) Each client $c$ uses the GHN to predict weights $\theta_c$ for its own specific architecture $f_c$, and updates the GHN weights $W_c$ locally using its own architecture and data (arched gray arrow from each client to itself); (3) Each client sends the locally updated GHN weights to the server (orange dashed lines); %
(4) Weight averaging is performed on the server (arched gray arrow from the GHN to itself);. %

\subsection{Method}\label{subsec:method}
\vspace{-5pt}
\paragraph{Representing neural architectures as graphs.}

A neural architecture can be represented in many different ways. Early chain-structured architectures such as VGG \cite{simonyan2014very} and AlexNet \cite{krizhevsky2012imagenet} can easily be represented as sequences of layers. Recently, several  architectures that have richer connectivity structures have been proposed~\cite{ronneberger2015u, targ2016resnet, huang2017densely}. Sequential representation is not suitable for these architectures. In contrast, all neural architectures are computational graphs, which makes a graph representation a much more natural choice. %
We use graph representations to ensure generality in architecture space.

\begin{figure}[t]
    \centering
    \includegraphics[width=0.75\textwidth]{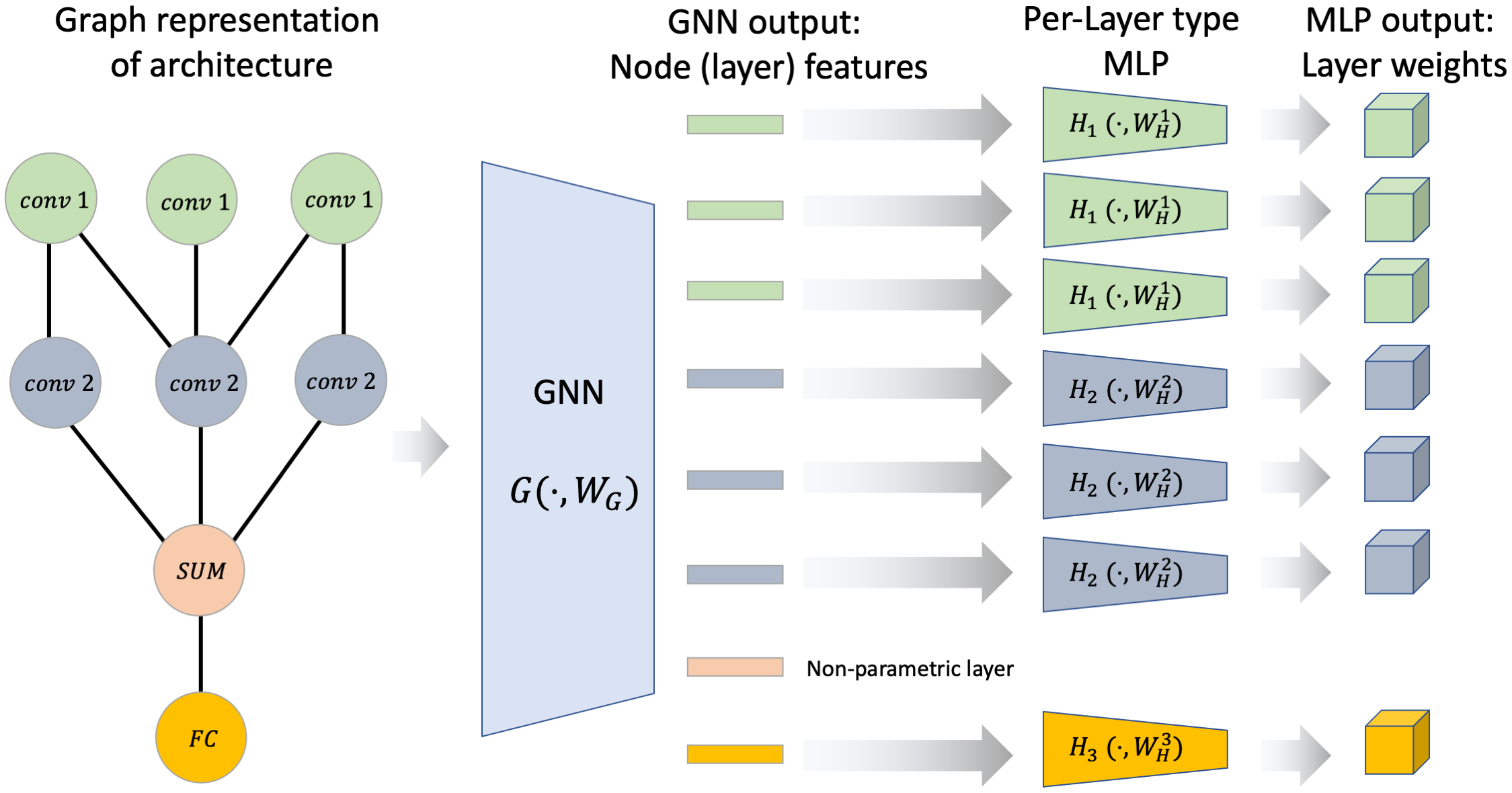}
    \caption{The GHN architecture. GHN is composed of two sub-networks: a Graph neural network (GNN) $G$ and a set of MLPs $\{H_l\}_{l\in L}$, one for each layer type. We input a graph representation of an architecture to the GHN, with the colored node features corresponding to different layer types. We then apply the GNN to produce node features. Finally, the node features for the parametric layers are further processed by an MLP $H_l$ to attain the final layer weights.}
    \label{fig:ghn}
\end{figure}

Given a neural architecture, we represent it as a graph  $\mathcal{A} = (\mathcal{V}, \mathcal{E}, \mathcal{X})$ in the following way. The set of vertices $\mathcal{V}$ contains a vertex $v$ for each parametric layer in the architecture. For example, a convolutional layer with weights of dimensions $3\times 3\times 64\times 128$. \rev{In a similar way, we use another type of nodes to represent non-parametric operation in the network, such as summation or concatenation of matrices, which are frequently used in ResNet-like architectures. See example in Figure \ref{fig:ghn}}.
The set of edges $\mathcal{E}$ represents the computational flow of the architecture: there is an edge between the nodes $v$ and $u$ (i.e., $e=(v,u)\in \mathcal{E}$) if the output of the layer represented by $v$ is the input of a layer represented by $u$. $\mathcal{X}\in \mathbb{R}^{|\mathcal{V}|\times k}$ is a  matrix that holds the input node features. Initially, each node is equipped with categorical (one-hot) features indicating the layer type they represent, denoted by $L=\{l_1,\dots,l_k\}$. Each categorical layer type specifies the following: linear/conv, stride, kernel size, activation and feature dimensions.  As an example, the convolution layer discussed above may be represented as $l_1$. 
\rev{According to the desired architecture family, the nodes in the graph can represent different computational blocks of different granularity. Granularity can range from a single layer to complex blocks, so nodes can even represent complex mechanisms like attention. Throughout this paper, we consider the ResNets architecture family and model layers as nodes.}

\paragraph{GHN Architecture.} Originally introduced for neural architecture search, a GHN~\citep{zhang2020graph,knyazev2021parameter} is a generalization of standard hypernetworks that allows generating weights for heterogeneous network architectures. The GHN weight prediction can be broken down into two stages: (1) Processing a graph representation of an architecture using a GNN and then (2) Predicting weights for each layer in that architecture. We now explain these two stages in detail.

In the first stage, our hypernetwork takes as input a graph representation $\mathcal{A}$ of an architecture  and processes it using a  $T$-layer graph neural network $G(\cdot;W_G)$ with learnable parameters $W_G$ (we use $T=6$ in our implementation). This process outputs latent representations $h^T_v$ for each node $\{v\in\mathcal{V}\}$. We use maximally expressive message passing GNN layers, as introduced in \cite{morris2019weisfeiler}. These layers have the following form:
\begin{equation}
\small
    h_v^{(t)} = \sigma\left( A^{(t)}h_v^{(t-1)} + B^{(t)}\sum_{\{u\mid  (u,v)\in \mathcal{E}\}} h_u^{(t-1)} + b^{(t)}\right).
\end{equation}
Here, $t \in \{1,\dots,T\}$ represents the depth of the layer, $A^{(t)},B^{(t)},b^{(t)}$ are layer-specific parameters. and $\sigma$ is a non-linear activation function such as ReLU. As mentioned above, we denote the concatenation of the parameters $A^{(t)},B^{(t)},b^{(t)},~t\in \{1,...,T \}$ as $W_G$. As shown in several recent works, the node features extracted by such GNNs are a representation of the local neighbourhood around each node \citep{xu2018powerful,morris2019weisfeiler,yehudai2021local}.

At the second stage, we use a set of MLPs $\big\{H_l(\cdot;W^l_H)\big\}_{l\in L}$ with learnable parameters $W^l_H,~l\in L$ to map latent node representations $h_v^T$ to layer weights $\theta^v$: 
\begin{equation}
    \theta^v = H_{l(v)}(h_v^T;W^{l(v)}_H),\quad v \in \mathcal{V}, 
\end{equation}
where $l(v)$ is a categorical variable that determines the type of the layer represented by the node~$v$. We note that a single MLP will not suffice since there are multiple layer types with different output sizes. We denote the concatenation of $\big\{W^{l}_H\big\}_{l\in L}$ as $W_H$.  For a particular client $c$, the weights $\{\theta^v \}_{v\in \mathcal{V}}$ are concatenated to form the client's weight vector $\theta_c(\mathcal{A}_c;W_G,W_H)$ mentioned above.

\paragraph{Objective and training procedure.} 
Based on the GHN formulation, we can now state our training objective: we look for optimal GHN paramters $(W^*_G,W^*_H)$ that simultaneously minimize the empirical risk of all clients:
\begin{equation}
    (W^*_G,W^*_H) = \text{argmin}_{(W_G,W_H)} \sum_{c=1}^C \sum_{j=1}^{n_c} \ell(f_c(x_{cj};\theta_c(\mathcal{A}_c;W_G,W_H)),y_{cj}). 
\end{equation}

The training procedure of our GHN is based on local updates of the GHN weights, performed by all clients, followed by a GHN weight averaging process on the shared server. More specifically, the local optimization at each client $c$ aims to solve the following client-specific minimization problem:
\begin{equation} \label{eq:localupdate}
        (W^{c*}_G,W^{c*}_H) = \text{argmin}_{(W_G,W_H)} \sum_{j=1}^{n_c} \ell(f_c(x_{cj};\theta_c(\mathcal{A}_c;W_G,W_H)),y_{cj}), 
\end{equation}
and performs a predefined number of SGD %
iterations (or similar gradient-based optimization method). Locally updated weights are then averaged at the server side and redistributed to the clients for further updates. 
The procedure is summarized in Algorithm \ref{alg:training}.

\rev{We note that FedAvg can be seen as a specific instance of \ShortName-GHN. To see that, consider the standard FL setting where all client architectures are the same and the GNN implements the identity map. Consequently, all clients will have the same latent node representations $h^0_v$: the 1-hot encoding of the layer. If the MLPs $\big\{H_l(\cdot;W^l_H)\big\}_{l\in L}$ are all implemented as a linear mapping (i.e., $\theta^v = H_{l(v)}(h_v) = W_{l(v)}h_v$) then averaging their parameters $W_{l(v)}$ is equivalent to directly averaging the client network parameters.}

\begin{algorithm}[t] \label{alg:training}
\footnotesize
\SetAlgoLined
\KwIn{R: number of communication rounds, C: number of clients, L: local updates.}
 Initialize shared GHN weights %
 \;
 \For{r = 1,...,R}{
 Server shares current GHN weights $(W_G,W_H$) with all clients $c\in \{1,...,C\}$ \;
 \For{c = 1,...,C}{
  Update GHN weights by local optimization for L update steps on client $c$ (see Eq.~\eqref{eq:localupdate})\;
  Send updated GHN weights $(W^c_G,W^c_H)$ to the server \;
  }
  Average GHN weights: $W_G \leftarrow \frac{1}{C}\sum W^c_G$, $W_H \leftarrow \frac{1}{C}\sum W^c_H$ \; 
 }
 \caption{\ShortName-GHN}
\end{algorithm}

\vspace{-5pt}
\paragraph{Hypernetwork weight initialization.}\label{hnet_weight_init}
Poor weight initialization of CNNs can have a detrimental effect to the training behavior causing either vanishing or exploding gradients. This has been carefully studied in \cite{glorot2010understanding,he2015delving}. When the network weights are outputs of another network, %
proper initialization of the hypernetwork parameters must be carried out so as to output the proper main network initialization. This has been observed by \cite{chang2019principled} who proposed a variance formula. We empirically found that initializing $W_H$ using zero bias and Xavier-normal initialization multiplied by  $\sqrt{2c_\text{in}/d_\text{lat}}$ translates to an accurate Kaiming initialization of the client network. Here $c_\text{in}$ and $d_\text{lat}$ are the number of channels in each  layer and the hidden layer dimension of $H_l$. See supplementary for more details.    
\vspace{-5pt}
\paragraph{Inference and local refinement.} At inference time, the trained GHN is applied to the graph representation of each architecture $\mathcal{A}$, yielding architecture-specific weights $\theta_c(\mathcal{A};W_G,W_H)$. We discuss the case of generalizing to architectures not seen during training in Section~\ref{subsec:generalization}. In contrast to the training stage, where the generated weights are used as-is, at the inference stage, they can be used as an initialization and then locally refined by using the client's local dataset. The amount of improvement depends on the amount of local data. Furthermore, since this data has already been used for training the shared GHN model, minor improvements should be expected before overfitting might occur. We found that a single refinement epoch of just the linear prediction head worked well in practice.

\subsection{Implementation details}\label{subsec:details}
\vspace{-5pt}
For full reproducibility of our results, we will release code upon publication. We ran an extensive hyperparameter search using \cite{wandb} for a 4 architecture setup using a fixed number of 500 epochs. 
We found the optimal values (which were used in all our experiments) to be: the GNN introduced in \citep{morris2019weisfeiler} with $T=6$ layers, a latent dimension of $51$, SGD with a learning rate of $0.009$ and cosine scheduling. Inspired by \cite{zhang2020graph}, we experimented with both synchronous and asynchronous message-passing that respects the directional graph structure, yet experiment did not indicate an advantage to the latter. For the client architectures, we define $k=10$ different layer types. The different architectures and non-parametric layers are described in detail in the supplementary material. For each layer type, a 2-layer MLP hypernetwork is used with a latent dimension of $16$, and a leaky-ReLU activation function. We adopted the idea from \cite{littwin2019deep} and separately predict the scale $s$ and weight $W$ for each node so that the layer parameters become $sW$. Implementation was done in PyTorch~\cite{paszke2017automatic} using PyTorch Geometric~\citep{Fey/Lenssen/2019} and training was done on a cluster with NVIDIA DGX V100 GPUs. The range of parameter sweep and implementation details for baselines are provided in the supplementary.

\section{Experiments}\label{sec:exp}
Here we describe a comparative experimental study of our approach. In Section~(\ref{subsec:comparison}) we review  the methods that we compare with. Sections~\ref{subsec:cifar} \& \ref{subsec:xray} provide results for two applications. Finally, we show ablations including generalization to unseen architectures~(Sec.~\ref{subsec:generalization}), the effect of communication rate~(Sec.~\ref{subsec:comm_rate}), and the importance of the GNN~(Sec.~\ref{subsec:no_sharing}).  
In the supplementary material, we also demonstrate strong results with our method in unbalanced data distribution scenarios.

We evaluate \ShortName-GHN performance in two tasks. For natural image classification, we use two standard benchmarks in FL: CIFAR-10 and CIFAR-100. Another task that emphasizes the importance of enabling \ShortName between medical facilities involves disease classification in chest x-ray scans. CNNs are widely adopted for image classification tasks, and architectures mostly differ in depth, layer types and graph connectivity (e.g. residual connection). We thus ran each experiment with four different network architectures, named Arch 0-3. Arch 0 is a standard ResNet18~\cite{he2015deep}, Arch 1 has 10 layers and no skip connections, Arch 2 has 12 layers and skip connections are used only in the \textit{first} 8 layers, and Arch 3 has 12 (different) layers with skip connection between the \textit{last} 8 layers. Detailed designs can be found in Figure~\ref{fig:archs} in the supplementary. In all experiments, we split the data (either uniformly or following a Dirichlet distributions, as indicated in the setup) between 4, 8, and 16 clients while maintaining an even distribution of the different architectures among clients (for instance, in the 8 clients case, we use 2 clients for each architecture). 

\subsection{Compared baselines}\label{subsec:comparison}%
Since we are the first to address the \ShortName setup (FL between substantially different architectures that are kept undisclosed, and does not require  external data), we compare our performance with two variants of previous methods that we adapted to the \ShortName setup. (1) A local-distillation baseline, which we name \textbf{\ShortName-distillation,} a variant of the distillation technique proposed by \cite{lin2020ensemble}. This baseline is based on locally available data and does not require the clients to disclose their architectures. In detail, for the first step we use standard federated averaging \citep{mcmahan2017learning} to train a shared architecture. The trained model is then sent to the client for distillation using only local data. (2) A new variant of \textbf{pFedHN} \citep{shamsian2021personalized}, where heads of individual client models where modified to account for different architectures (different number of outputs according to the number of parameters). 

In addition to these baselines, we also report results of \textbf{Local training} where each client performs standard (non-FL) training on its local train samples. This can be seen as a lower bound. For completion, we also include an \textbf{Upper bound} score. This is the result if all the clients used the most expressive Arch 0, and trained in a standard FL. 

\subsection{Results on CIFAR-10 and CIFAR-100}\label{subsec:cifar}
We experiment with the widely known CIFAR-10/CIFAR-100 \citep{cifar} image classification datasets that contain 60,000 $32\times 32$ natural color images in 10/100 classes with predefined train/test split. The results under \ShortName are summarized in Table~\ref{tab:cifar}. Our method outperforms the local distillation \rev{ and pFedHN baselines} consistently except on CIFAR-10 when local data is $25\%$ where we perform similarly or slightly worse than local distillation.  
Importantly, as can be expected, as the amount of locally available data decreases, local-distillation deteriorates rapidly. By contrast, our proposed method shows gradual degradation, resulting in an improvement of about 20 points in average accuracy over the baseline when local data percentage is at $6.25\%$. \rev{Evidently, despite being tuned to its best performing parameters per setup (dataset+split) pFedHN performs poorly. This can be attributed to weight sharing only occurring at the shared MLP responsible for the encoding the architecture, so the separate predictions heads do not benefit from federation. A reduced version of our method without the graph structure is also examined. This no-graph variant, which can be seen as a set network without information about the connectivity between the model layers, already performs very well. The importance of the graph is discussed in detail in Section~\ref{subsec:no_sharing}.
\begin{table}[H]
\scriptsize
  \centering
  \caption{Accuracy on CIFAR-10/100 datasets }
    
    \resizebox{0.99\textwidth}{!}{
    \addtolength{\tabcolsep}{-2pt}
    \begin{tabular}{cl|ccccc|ccccc}
    \toprule
          &       & \multicolumn{5}{c|}{CIFAR-100} & \multicolumn{5}{c}{CIFAR-10} \\
          
          data \% & method & \multicolumn{1}{l}{Arch 0} & \multicolumn{1}{l}{Arch 1} & \multicolumn{1}{l}{Arch 2} & \multicolumn{1}{l}{Arch 3} &
          \multirow{2}[1]{*}{Avg.} &
          \multicolumn{1}{l}{Arch 0} & \multicolumn{1}{l}{Arch 1} & \multicolumn{1}{l}{Arch 2} & \multicolumn{1}{l}{Arch 3} & \multirow{2}[1]{*}{Avg.} \\

     {}&  & \multicolumn{1}{l}{Original} & \multicolumn{1}{l}{No Skip} & \multicolumn{1}{l}{Skip first} & \multicolumn{1}{l}{Skip last} &
     {} &\multicolumn{1}{l}{Original} & \multicolumn{1}{l}{No skip } & \multicolumn{1}{l}{Skip first} & \multicolumn{1}{l}{Skip last} & {} \\
     
    \midrule
    \multirow{5}[2]{*}{25\%} & Upper-bound & \multicolumn{4}{c}{73.2} & {} & \multicolumn{4}{c}{93.6} & {}
    \\ & Local training & 49.5 & 50.2 & \textbf{52.2} & 48.2 & 50.1 & 86.3 & 84.6 & 84.7 & 85.4 & 85.3 \\
    & pFedHN & 38.0 & 38.7 & 46.5 & 34.7 & 39.5 & 85.8
    & 84.1 & 85.8 & 84.4 & 85.0 \\
          & \ShortName Distillation & 51.1 & 44.8 & 49.0 & 46.8 & 47.9 & \textbf{90.0} & \textbf{89.3} & \textbf{89.4} & \textbf{90.1} & \textbf{89.7} \\
          & Ours  & \textbf{56.3} & 51.8 & 50.1 & 51.9 & 52.5 & 89.9 & 87.4  & 86.0 & 89.0 & 88.1 \\
          & Ours (no-graph)  & 55.5 & \textbf{52.7} & 49.5 & \textbf{55.2} & \textbf{53.2} & 89.9 & 87.4  & 86.0 & 89.0 & 88.1 \\
    \midrule
    \multirow{5}[2]{*}{12.50\%} & Upper-bound & \multicolumn{4}{c}{71.1} & {} & \multicolumn{4}{c}{93.2} & {}
    \\ & Local training & 36.4 & 32.5 & 38.1 & 33.5 & 35.1 & 79.2  & 75.6 & 74.5 & 80.0 & 77.3 \\
    & pFedHN & 17.8 & 25.0 & 24.0 & 20.2 & 21.8 & 77.6
    & 76.0 & 78.3 & 74.8 & 76.7 \\
          & \ShortName Distillation & 44.2 & 33.2 & 41.3 & 35.2 & 38.5 & 85.7 & 84.6 & 84.2 & 83.3 & 84.5 \\
          & Ours & \textbf{52.3} & \textbf{49.9} & \textbf{48.4} & 49.9 & \textbf{50.1} & \textbf{88.9} & 85.9 & \textbf{85.8} & 87.0 & \textbf{86.9} \\
          & Ours (no-graph)  & 51.5 & 48.7 & 45.7 & \textbf{50.4} & 49.1 & 88.3 & \textbf{86.3} & 84.9 & \textbf{87.4} & 86.7 \\
    \midrule
    \multirow{5}[2]{*}{6.25\%} & Upper-bound & \multicolumn{4}{c}{71.3} & {} & \multicolumn{4}{c}{92.9} & {} 
    \\ & Local training & 22.7 & 19.4 & 22.0  & 22.0  & 21.5 & 76.9 & 75.6 & 75.6 & 74.5 & 75.7 \\
    & pFedHN & 13.6 & 14.4 & 16.4 & 12.6 & 14.3 & 58.2 & 56.6 & 57.3 & 50.7 & 55.7 \\
          & \ShortName Distillation  & 25.3 & 23.1 & 30.0 & 23.6 & 25.5 & 77.4 & 75.9 & 75.7 & 74.8 & 76.0 \\
          & Ours  & 47.3 & \textbf{46.6} & \textbf{44.0} & \textbf{47.0} & \textbf{46.2} & 86.4 & 83.8 & \textbf{83.0} & 85.3 & 84.6 \\
          & Ours (no-graph)  & \textbf{47.8} & 46.4 & 41.2 & 46.9 & 45.6 & \textbf{86.9} & \textbf{84.5} & 82.7 & \textbf{86.2} & \textbf{85.1} \\
    \bottomrule
    \end{tabular}%
    }
  \label{tab:cifar}%
  \vspace{-5pt}
\end{table}%
}

\subsection{Results on Chest X-ray}\label{subsec:xray}
\vspace{-5pt}
As discussed in the introduction, a main motivation for \ShortName is to allow cross-organizational collaborations. This is especially important when multiple entities have access to valuable data that cannot be shared, but use different neural architectures for processing it. Such is the case for medical clinics and hospitals. To showcase the application of our method to such a real-world problem, we tested it using medical data of X-ray images. The Chest X-ray~\cite{wang2017chestxray} dataset comprises 112,120 frontal-view X-ray images of 30,805 unique patients with fourteen disease image labels (images can have multiple labels), extracted from medical reports using natural language processing. %
Table~\ref{tab:nih} reports the average AUC score across 14 binary classification tasks for our method as well as the baselines. Our method achieves consistently superior performance. 
\begin{table}[H]
\scriptsize

  \centering
  \caption{Average AUC on NIH Chest X-ray dataset.}
    \begin{tabular}{cl|ccccc}
          & \multicolumn{1}{r}{} &       &       &       &  & \\
    \midrule
          &       & \multicolumn{5}{c}{ChestX-ray~\citep{wang2017chestxray}} \\
    \midrule
    data \% & method & \multicolumn{1}{l}{Arch 0} & \multicolumn{1}{l}{Arch 1} & \multicolumn{1}{l}{Arch 2} & \multicolumn{1}{l}{Arch 3} & \multirow{2}[1]{*}{Avg.} \\
    & &\multicolumn{1}{l}{Original} & \multicolumn{1}{l}{No Skip 2} & \multicolumn{1}{l}{Skip first} & \multicolumn{1}{l}{Skip last} & \\
    \midrule
    \multirow{5}[2]{*}{25\%} & Upper-bound & \multicolumn{4}{c}{78.0} &   
    \\ & Local training & 60.5 & 61.8  & 58.7 & 60.3 & 60.3\\
    & pFedHN & 62.4  & 60.5 & 63.0 & 63.7 & 62.4 \\
          & \ShortName Distillation & 67.5 & 67.8 & 65.4 & 69.8 & 67.6 \\
          & Ours& \textbf{72.2} & \textbf{73.2} & \textbf{70.0} & \textbf{70.5} & \textbf{71.5}\\
    \midrule
    \multirow{5}[2]{*}{12.50\%} & Upper-bound & \multicolumn{4}{c}{78.0} &  
    \\  & Local training & 59.5  & 59.4 & 57.1 & 59.8 & 59.0 \\
    & pFedHN & 62.5  & 59.9 & 58.4 & 61.0 & 60.5 \\
          & \ShortName Distillation & 66.4 & 64.0 & 62.9 & 67.6 & 65.2 \\
          & Ours  & \textbf{74.0} & \textbf{71.9} & \textbf{70.6} & \textbf{73.5} & \textbf{72.5} \\
    \midrule
    \multirow{5}[2]{*}{6.25\%} & Upper-bound & \multicolumn{4}{c}{77.0} &  
    \\ & Local training & 59.2 & 58.5 & 56.9 & 58.9 & 58.4 \\
    & pFedHN & 60.4 & 56.3 & 58.0 & 60.0 & 58.7 \\
          & \ShortName Distillation  & 65.7 & \textbf{62.8} & 61.72 & 66.5 & 64.2 \\
          & Ours  & \textbf{69.4} & 62.4 & \textbf{64.4} & \textbf{68.0} & \textbf{66.1} \\
    \bottomrule
    \end{tabular}%
  \label{tab:nih}%
\end{table}%

\subsection{Generalization to unseen architectures}\label{subsec:generalization}

In standard FL, a client that did not participate in the training procedure may still benefit from the pre-trained network as long as its data is similar enough to that used in training \cite{shamsian2021personalized}. 
We want \ShortName-GHN to enjoy a similar generalization, namely, that the GHN could generalize to clients whose architecture has not been observed during training. 
Since neural architectures often share local structures, as with the various variants %
ResNets, when a new architecture is composed of local structures that have previously been seen, it may be feasible to generalize to a new composition of these known components (compositional generalization). The reason is that $T$-layer message-passing GNNs essentially encode the $T$-hop local connectivity pattern around each node. When the same $T$-hop neighborhood appears in a new architecture, possibly in a different position in the computational graph, the GNN can predict reasonable weights since it was trained on such local structures \citep{yehudai2021local}. %

\begin{wrapfigure}[15]{r}{0.45\textwidth}
  \begin{center}
  \vspace{-15pt}
    \includegraphics[width=\textwidth]{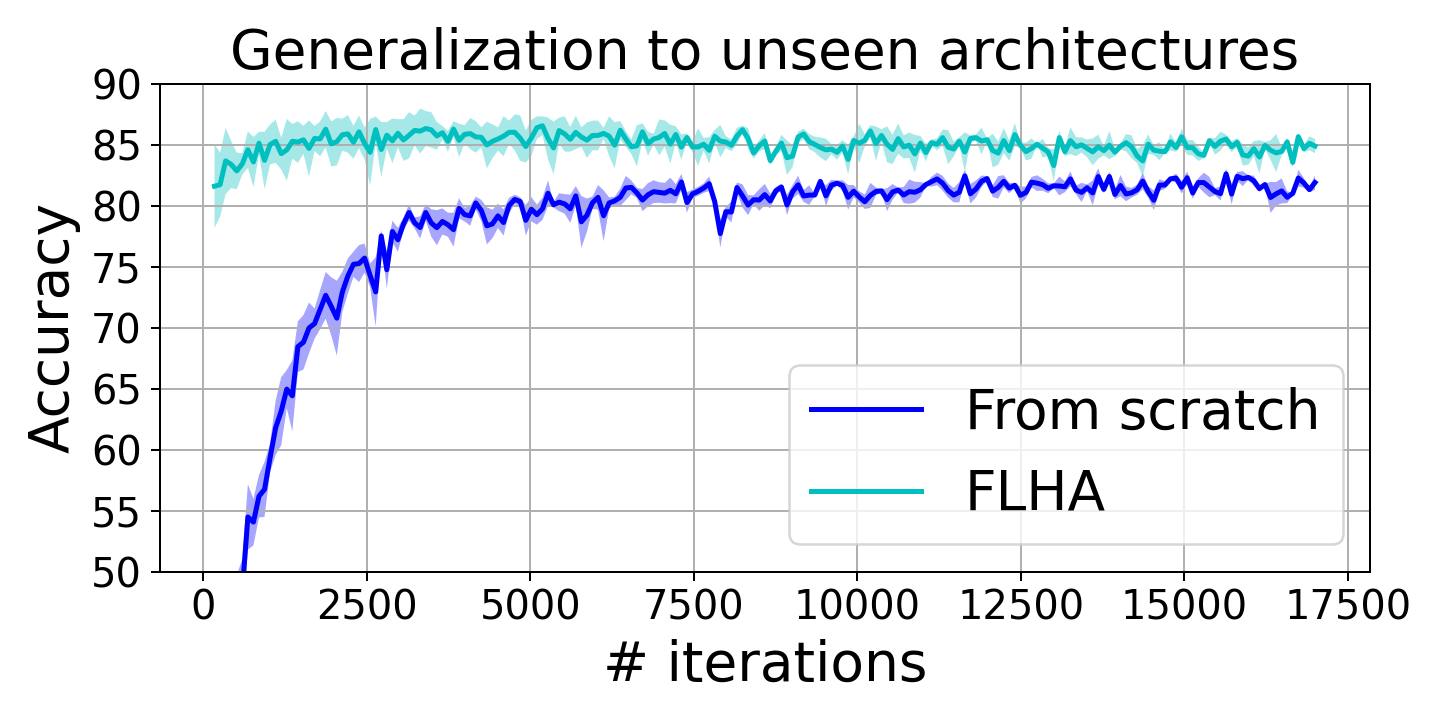}
  \end{center}
  \vspace{-10pt}
  \caption{Generalization to unseen architectures: our method (light blue) quickly ramps up to high performance and maintains a considerable gap compared to training from scratch (dark blue). %
  }
  \label{fig:generalization}
\end{wrapfigure} 

A clear benefit of such generalization is that given a new architecture, our GHN can immediately populate it to give a better initialization. To test this we ran a ``leave one out'' experiment using CIFAR-10 dataset, where we let 3 clients with 3 different architectures train in an \ShortName fashion. We then introduce a 4th architecture and refine it using only local data. We compare against training that architecture from scratch on that same data. The results shown in Figure~\ref{fig:generalization} and Table \ref{tab:gen_arch} in the supplementary are very encouraging. When refining on local data, the unseen architecture performance ramps up very quickly and reaches similar performance to that achieved by the architectures that participated in the \ShortName-GHN process. In contrast, the training-from scratch alternative takes a much longer to converge and reaches a consistently lower final performance. 
\rev{We further stress tested the generalization capabilities, by introducing a much smaller CNN with only 4 layers. A practical scenario of this sort is training a model on an edge device together with stronger models on the server. Table \ref{tab:small_arch} in the supplementary shows the performance of the 4 architectures used in our main experiments, and how they are influenced when each is replaced by a smaller architecture. The average drop in performance of $2.2 \pm 1.4$ keeps the performance well above the local-training alternative. Despite the different architecture, when initialized with a model trained on the four main architectures, the fast convergence property persist (see Figure~\ref{fig:small_arch}).}

\subsection{Communication rate}\label{subsec:comm_rate}

\begin{wrapfigure}[10]{r}{0.40\textwidth}
  \begin{center}
  \vspace{-20pt}
    \includegraphics[width=\textwidth]{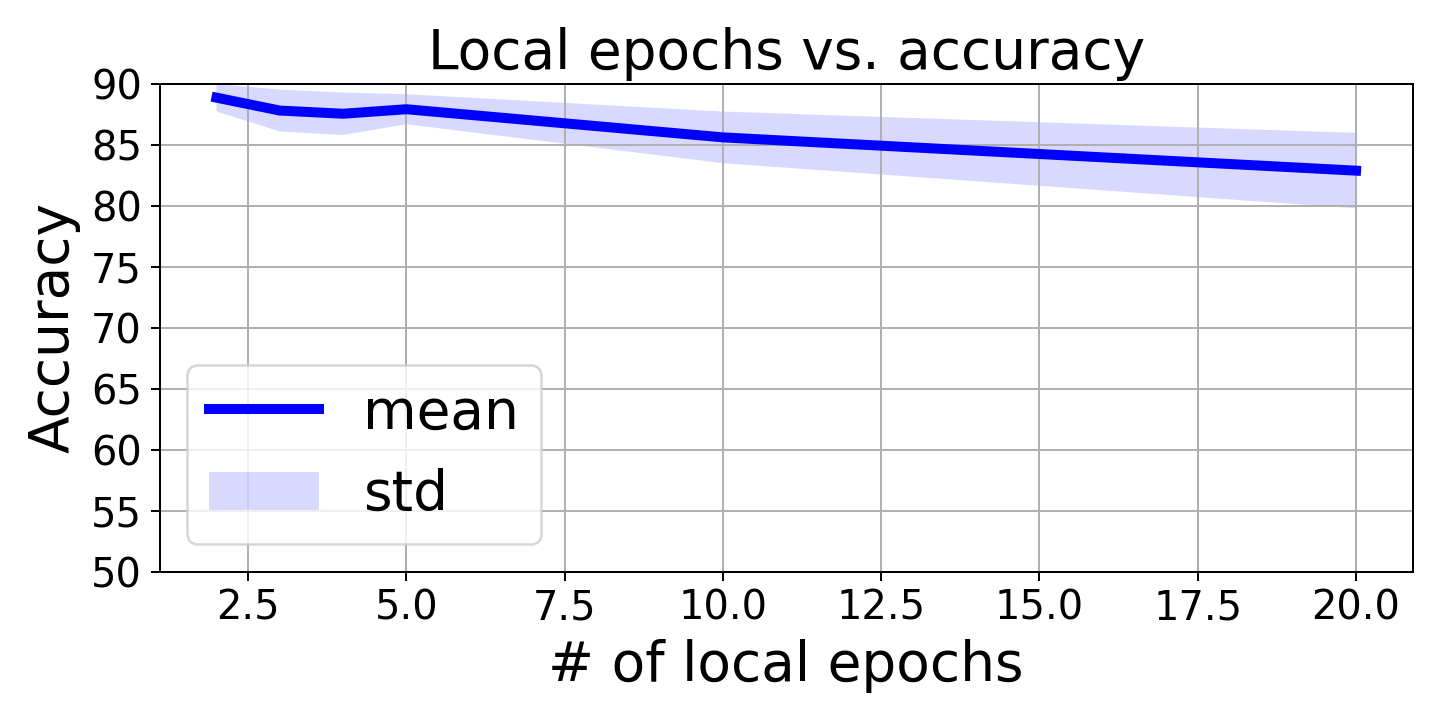}
  \end{center}
  \vspace{-15pt}
  \caption{\ShortName-GHN on CIFAR-10 with different communication rates.}
  \label{fig:comm}
\end{wrapfigure}
In FL, the amount of data transfer could become a bottleneck. A common remedy is to reduce the frequency of clients-server communication %
To study the effect of communication rates on the \ShortName-GHN performance, we trained 4 architectures on CIFAR-10. At each experiment, clients train locally for $L$ epochs before sending local GHN weights to the server for averaging. Client-server communication occurs $R$ times, keeping $L\times R$ at a constant value of $1000$. Figure~\ref{fig:comm} shows the results for R between 1 and 20. While more frequent communication leads to better performance, we observe that even with 5 times less frequent communication the overall performance decreases by less than $1\%$.  

\subsection{The importance of encoding the graph structure}\label{subsec:no_sharing}
\vspace{-5pt}
\rev{Our \ShortName-GHN uses the same layer types for all models. As these layers are repeated within the computational graphs of the different models, more sharing can be done, leading to a more efficient use of the federation.
 Importantly, while same layers could be directly averaged, it is important that their location within the graph is encoded. This is done via message passing in our GNN. To verify its importance, we implemented a variant of our method that does not use a GNN and instead directly averages MLPs corresponding to the same layer types. Table~\ref{tab:gnn} shows this ablation, for which we used CNNs with 4,7,10 and 13 layers all of type $3\times 3\times 64\times 64$ except for the first and last. With this amount of sharing, the results highlight that the GNN is instrumental for performance gains.}

\begin{table}[htbp]
\scriptsize
  \centering
  \caption{\textbf{The importance of the GNN.} Under an \ShortName setup we use 4 architectures sharing a repeated layer type. The result show significant gains due to message passing informing the layer encoding with their placement within the architecture.}
    \begin{tabular}{cc|cccc|rrrr}
    \toprule
    \multirow{2}[1]{*}{data \%} & \multirow{2}[1]{*}{GNN} & \multicolumn{4}{c|}{CIFAR10}  & \multicolumn{4}{c}{CIFAR100} \\
          &       & 4 layers & 7 layers & 10 layers & 13 layers & \multicolumn{1}{c}{4 layers} & \multicolumn{1}{c}{7 layers} & \multicolumn{1}{c}{10 layers} & \multicolumn{1}{c}{13 layers} \\
          \midrule
    \multirow{2}[1]{*}{25\%} & \xmark & 55.3  & 75.5  & 75.1  & 69.9  & 27.8  & 40.0  & 36.5  & 29.4 \\
          & \cmark & 63.6  & 77.4  & 77.0  & 74.4  & 26.4  & 41.1  & 40.4  & 33.5 \\
    \midrule
    \multirow{2}[2]{*}{12.50\%} & \xmark & 60.2  & 75.5  & 73.8  & 70.1  & 32.4  & 41.5  & 41.6  & 36.8 \\
          & \cmark     & 67.7  & 78.2  & 77.8  & 74.4  & 36.1  & 43.5  & 42.3  & 35.0 \\
    \bottomrule
    \end{tabular}%
  \label{tab:gnn}%
\end{table}%

\vspace{-5pt}

\section{Conclusion and limitations}
\vspace{-5pt}
We proposed a new setup for federated learning called \ShortName, and a first solution to this problem based on a graph hypernetwork. Our experiments yielded positive results: by modeling neural networks as graphs with  layers as nodes, we have shown that GNNs can utilize recurring structures and facilitate efficient federation during learning and even generalization to new architectures.%
There are, however, a few limitations to our solution. First, it is limited to architectures built from predefined building blocks. In theory, this set could be arbitrarily large, but architectures that use disjoint sets of nodes may lose efficiency when sharing knowledge. Future work might benefit from combining the inclusion structure proposed on ~\cite{diao2021heterofl} to group layers that only differ in the number of channels. Second, as indicated by the upper-bound performance, our solution still is not yet comparable with a same-architecture FL performance. Finally, a more robust aggregation technique would alleviate the undesirable degradation of performance when local data is limited. Our hope is that this research will lead to further study of this important setup.

\textbf{Acknowledgement:}
We thank the following people for useful discussions and proofreading: Leonidas Guibas, Zan Gojcic, Francis Williams, and Cinjon Resnick.

\bibliography{egbib}

\begin{thebibliography}{54}
\providecommand{\natexlab}[1]{#1}
\providecommand{\url}[1]{\texttt{#1}}
\expandafter\ifx\csname urlstyle\endcsname\relax
  \providecommand{\doi}[1]{doi: #1}\else
  \providecommand{\doi}{doi: \begingroup \urlstyle{rm}\Url}\fi

\bibitem[Agarwal et~al.(2018)Agarwal, Suresh, Yu, Kumar, and
  McMahan]{agarwal2018cpsgd}
Naman Agarwal, Ananda~Theertha Suresh, Felix Xinnan~X Yu, Sanjiv Kumar, and
  Brendan McMahan.
\newblock cpsgd: Communication-efficient and differentially-private distributed
  sgd.
\newblock In S.~Bengio, H.~Wallach, H.~Larochelle, K.~Grauman, N.~Cesa-Bianchi,
  and R.~Garnett (eds.), \emph{Advances in Neural Information Processing
  Systems}, volume~31. Curran Associates, Inc., 2018.
\newblock URL
  \url{https://proceedings.neurips.cc/paper/2018/file/21ce689121e39821d07d04faab328370-Paper.pdf}.

\bibitem[Biewald(2020)]{wandb}
Lukas Biewald.
\newblock Experiment tracking with weights and biases, 2020.
\newblock URL \url{https://www.wandb.com/}.
\newblock Software available from wandb.com.

\bibitem[Brock et~al.(2018)Brock, Lim, Ritchie, and Weston]{brock2017smash}
Andrew Brock, Theo Lim, J.M. Ritchie, and Nick Weston.
\newblock {SMASH}: One-shot model architecture search through hypernetworks.
\newblock In \emph{International Conference on Learning Representations}, 2018.
\newblock URL \url{https://openreview.net/forum?id=rydeCEhs-}.

\bibitem[Chang et~al.(2019)Chang, Flokas, and Lipson]{chang2019principled}
Oscar Chang, Lampros Flokas, and Hod Lipson.
\newblock Principled weight initialization for hypernetworks.
\newblock In \emph{International Conference on Learning Representations}, 2019.

\bibitem[Chen et~al.(2021)Chen, Shlezinger, Poor, Eldar, and
  Cui]{chen2021communication}
Mingzhe Chen, Nir Shlezinger, H~Vincent Poor, Yonina~C Eldar, and Shuguang Cui.
\newblock Communication-efficient federated learning.
\newblock \emph{Proceedings of the National Academy of Sciences}, 118\penalty0
  (17), 2021.

\bibitem[Dai et~al.(2019)Dai, Yan, Zhou, Yang, Ng, Cheng, and
  Fan]{dai2019hyper}
Xinyan Dai, Xiao Yan, Kaiwen Zhou, Han Yang, Kelvin Kai~Wing Ng, James Cheng,
  and Yu~Fan.
\newblock Hyper-sphere quantization: Communication-efficient {SGD} for
  federated learning.
\newblock \emph{CoRR}, abs/1911.04655, 2019.
\newblock URL \url{http://arxiv.org/abs/1911.04655}.

\bibitem[Diao et~al.(2021)Diao, Ding, and Tarokh]{diao2021heterofl}
Enmao Diao, Jie Ding, and Vahid Tarokh.
\newblock Hetero{\{}fl{\}}: Computation and communication efficient federated
  learning for heterogeneous clients.
\newblock In \emph{International Conference on Learning Representations}, 2021.
\newblock URL \url{https://openreview.net/forum?id=TNkPBBYFkXg}.

\bibitem[Fey \& Lenssen(2019)Fey and Lenssen]{Fey/Lenssen/2019}
Matthias Fey and Jan~E. Lenssen.
\newblock Fast graph representation learning with {PyTorch Geometric}.
\newblock In \emph{ICLR Workshop on Representation Learning on Graphs and
  Manifolds}, 2019.

\bibitem[Glorot \& Bengio(2010)Glorot and Bengio]{glorot2010understanding}
Xavier Glorot and Yoshua Bengio.
\newblock Understanding the difficulty of training deep feedforward neural
  networks.
\newblock In \emph{Proceedings of the thirteenth international conference on
  artificial intelligence and statistics}, pp.\  249--256. JMLR Workshop and
  Conference Proceedings, 2010.

\bibitem[Ha et~al.(2017)Ha, Dai, and Le]{ha2016hypernetworks}
David Ha, Andrew~M. Dai, and Quoc~V. Le.
\newblock Hypernetworks.
\newblock In \emph{5th International Conference on Learning Representations,
  {ICLR} 2017, Toulon, France, April 24-26, 2017, Conference Track
  Proceedings}. OpenReview.net, 2017.
\newblock URL \url{https://openreview.net/forum?id=rkpACe1lx}.

\bibitem[Hamilton et~al.(2018)Hamilton, Ying, and
  Leskovec]{hamilton2018inductive}
William~L. Hamilton, Rex Ying, and Jure Leskovec.
\newblock Inductive representation learning on large graphs, 2018.

\bibitem[Hanzely \& Richt{\'{a}}rik(2020)Hanzely and
  Richt{\'{a}}rik]{hanzely2020federated}
Filip Hanzely and Peter Richt{\'{a}}rik.
\newblock Federated learning of a mixture of global and local models.
\newblock \emph{CoRR}, abs/2002.05516, 2020.
\newblock URL \url{https://arxiv.org/abs/2002.05516}.

\bibitem[He et~al.(2015{\natexlab{a}})He, Zhang, Ren, and Sun]{he2015deep}
Kaiming He, Xiangyu Zhang, Shaoqing Ren, and Jian Sun.
\newblock Deep residual learning for image recognition, 2015{\natexlab{a}}.

\bibitem[He et~al.(2015{\natexlab{b}})He, Zhang, Ren, and Sun]{he2015delving}
Kaiming He, Xiangyu Zhang, Shaoqing Ren, and Jian Sun.
\newblock Delving deep into rectifiers: Surpassing human-level performance on
  imagenet classification, 2015{\natexlab{b}}.

\bibitem[Hinton et~al.(2015)Hinton, Vinyals, and Dean]{hinton2015distilling}
Geoffrey Hinton, Oriol Vinyals, and Jeff Dean.
\newblock Distilling the knowledge in a neural network, 2015.

\bibitem[Hsu et~al.(2019)Hsu, Qi, and Brown]{hsu2019measuring}
Tzu-Ming~Harry Hsu, Hang Qi, and Matthew Brown.
\newblock Measuring the effects of non-identical data distribution for
  federated visual classification, 2019.

\bibitem[Huang et~al.(2017)Huang, Liu, Van Der~Maaten, and
  Weinberger]{huang2017densely}
Gao Huang, Zhuang Liu, Laurens Van Der~Maaten, and Kilian~Q Weinberger.
\newblock Densely connected convolutional networks.
\newblock In \emph{Proceedings of the IEEE conference on computer vision and
  pattern recognition}, pp.\  4700--4708, 2017.

\bibitem[Kairouz et~al.(2021)Kairouz, McMahan, Avent, Bellet, Bennis, Bhagoji,
  Bonawitz, Charles, Cormode, Cummings, et~al.]{kairouz2019advances}
Peter Kairouz, H.~Brendan McMahan, Brendan Avent, Aur{\'{e}}lien Bellet, Mehdi
  Bennis, Arjun~Nitin Bhagoji, Kallista~A. Bonawitz, Zachary Charles, Graham
  Cormode, Rachel Cummings, et~al.
\newblock Advances and open problems in federated learning.
\newblock \emph{Found. Trends Mach. Learn.}, 14\penalty0 (1-2):\penalty0
  1--210, 2021.

\bibitem[Karimireddy et~al.(2020)Karimireddy, Kale, Mohri, Reddi, Stich, and
  Suresh]{karimireddy2020scaffold}
Sai~Praneeth Karimireddy, Satyen Kale, Mehryar Mohri, Sashank Reddi, Sebastian
  Stich, and Ananda~Theertha Suresh.
\newblock Scaffold: Stochastic controlled averaging for federated learning.
\newblock In \emph{International Conference on Machine Learning}, pp.\
  5132--5143. PMLR, 2020.

\bibitem[Kingma \& Ba(2015)Kingma and Ba]{kingma2014adam}
Diederik~P. Kingma and Jimmy Ba.
\newblock Adam: {A} method for stochastic optimization.
\newblock In Yoshua Bengio and Yann LeCun (eds.), \emph{3rd International
  Conference on Learning Representations, {ICLR} 2015, San Diego, CA, USA, May
  7-9, 2015, Conference Track Proceedings}, 2015.
\newblock URL \url{http://arxiv.org/abs/1412.6980}.

\bibitem[Klein et~al.(2015)Klein, Wolf, and Afek]{klein2015dynamic}
Benjamin Klein, Lior Wolf, and Yehuda Afek.
\newblock A dynamic convolutional layer for short range weather prediction.
\newblock In \emph{Proceedings of the IEEE Conference on Computer Vision and
  Pattern Recognition}, pp.\  4840--4848, 2015.

\bibitem[Knyazev et~al.(2021)Knyazev, Drozdzal, Taylor, and
  Romero-Soriano]{knyazev2021parameter}
Boris Knyazev, Michal Drozdzal, Graham~W Taylor, and Adriana Romero-Soriano.
\newblock Parameter prediction for unseen deep architectures.
\newblock In \emph{Advances in Neural Information Processing Systems}, 2021.

\bibitem[Konečný et~al.(2015)Konečný, McMahan, and
  Ramage]{konecny2015federated}
Jakub Konečný, Brendan McMahan, and Daniel Ramage.
\newblock Federated optimization:distributed optimization beyond the
  datacenter, 2015.

\bibitem[Konečný et~al.(2017)Konečný, McMahan, Yu, Richtárik, Suresh, and
  Bacon]{konecny2017federated}
Jakub Konečný, H.~Brendan McMahan, Felix~X. Yu, Peter Richtárik,
  Ananda~Theertha Suresh, and Dave Bacon.
\newblock Federated learning: Strategies for improving communication
  efficiency, 2017.

\bibitem[Krizhevsky et~al.(2009)Krizhevsky, Nair, and Hinton]{cifar}
Alex Krizhevsky, Vinod Nair, and Geoffrey Hinton.
\newblock {CIFAR-10} (canadian institute for advanced research), 2009.
\newblock URL \url{http://www.cs.toronto.edu/~kriz/cifar.html}.

\bibitem[Krizhevsky et~al.(2012)Krizhevsky, Sutskever, and
  Hinton]{krizhevsky2012imagenet}
Alex Krizhevsky, Ilya Sutskever, and Geoffrey~E Hinton.
\newblock Imagenet classification with deep convolutional neural networks.
\newblock \emph{Advances in neural information processing systems},
  25:\penalty0 1097--1105, 2012.

\bibitem[Kulkarni et~al.(2020)Kulkarni, Kulkarni, and Pant]{kulkarni2020survey}
Viraj Kulkarni, Milind Kulkarni, and Aniruddha Pant.
\newblock Survey of personalization techniques for federated learning.
\newblock In \emph{2020 Fourth World Conference on Smart Trends in Systems,
  Security and Sustainability (WorldS4)}, pp.\  794--797. IEEE, 2020.

\bibitem[Li et~al.(2020)Li, Sahu, Zaheer, Sanjabi, Talwalkar, and
  Smith]{sahu2018convergence}
Tian Li, Anit~Kumar Sahu, Manzil Zaheer, Maziar Sanjabi, Ameet Talwalkar, and
  Virginia Smith.
\newblock Federated optimization in heterogeneous networks.
\newblock In I.~Dhillon, D.~Papailiopoulos, and V.~Sze (eds.),
  \emph{Proceedings of Machine Learning and Systems}, volume~2, pp.\  429--450,
  2020.
\newblock URL
  \url{https://proceedings.mlsys.org/paper/2020/file/38af86134b65d0f10fe33d30dd76442e-Paper.pdf}.

\bibitem[Li et~al.(2019)Li, Milletar{\`\i}, Xu, Rieke, Hancox, Zhu, Baust,
  Cheng, Ourselin, Cardoso, et~al.]{li2019privacy}
Wenqi Li, Fausto Milletar{\`\i}, Daguang Xu, Nicola Rieke, Jonny Hancox, Wentao
  Zhu, Maximilian Baust, Yan Cheng, S{\'e}bastien Ourselin, M~Jorge Cardoso,
  et~al.
\newblock Privacy-preserving federated brain tumour segmentation.
\newblock In \emph{International Workshop on Machine Learning in Medical
  Imaging}, pp.\  133--141. Springer, 2019.

\bibitem[Li et~al.(2017)Li, Tarlow, Brockschmidt, and Zemel]{li2017gated}
Yujia Li, Daniel Tarlow, Marc Brockschmidt, and Richard Zemel.
\newblock Gated graph sequence neural networks, 2017.

\bibitem[Lin et~al.(2020)Lin, Kong, Stich, and Jaggi]{lin2020ensemble}
Tao Lin, Lingjing Kong, Sebastian~U. Stich, and Martin Jaggi.
\newblock Ensemble distillation for robust model fusion in federated learning,
  2020.

\bibitem[Littwin \& Wolf(2019)Littwin and Wolf]{littwin2019deep}
Gidi Littwin and Lior Wolf.
\newblock Deep meta functionals for shape representation.
\newblock In \emph{Proceedings of the IEEE/CVF International Conference on
  Computer Vision}, pp.\  1824--1833, 2019.

\bibitem[McMahan et~al.(2017)McMahan, Moore, Ramage, Hampson, and
  y~Arcas]{mcmahan2017communication}
Brendan McMahan, Eider Moore, Daniel Ramage, Seth Hampson, and Blaise~Aguera
  y~Arcas.
\newblock Communication-efficient learning of deep networks from decentralized
  data.
\newblock In \emph{Artificial Intelligence and Statistics}, pp.\  1273--1282.
  PMLR, 2017.

\bibitem[McMahan et~al.(2018)McMahan, Ramage, Talwar, and
  Zhang]{mcmahan2017learning}
H.~Brendan McMahan, Daniel Ramage, Kunal Talwar, and Li~Zhang.
\newblock Learning differentially private recurrent language models.
\newblock In \emph{International Conference on Learning Representations}, 2018.
\newblock URL \url{https://openreview.net/forum?id=BJ0hF1Z0b}.

\bibitem[Morris et~al.(2019)Morris, Ritzert, Fey, Hamilton, Lenssen, Rattan,
  and Grohe]{morris2019weisfeiler}
Christopher Morris, Martin Ritzert, Matthias Fey, William~L Hamilton, Jan~Eric
  Lenssen, Gaurav Rattan, and Martin Grohe.
\newblock Weisfeiler and leman go neural: Higher-order graph neural networks.
\newblock In \emph{Proceedings of the AAAI Conference on Artificial
  Intelligence}, volume~33, pp.\  4602--4609, 2019.

\bibitem[Mothukuri et~al.(2021)Mothukuri, Parizi, Pouriyeh, Huang,
  Dehghantanha, and Srivastava]{mothukuri2021survey}
Viraaji Mothukuri, Reza~M Parizi, Seyedamin Pouriyeh, Yan Huang, Ali
  Dehghantanha, and Gautam Srivastava.
\newblock A survey on security and privacy of federated learning.
\newblock \emph{Future Generation Computer Systems}, 115:\penalty0 619--640,
  2021.

\bibitem[Nachmani \& Wolf(2020)Nachmani and Wolf]{nachmani2020molecule}
Eliya Nachmani and Lior Wolf.
\newblock Molecule property prediction and classification with graph
  hypernetworks.
\newblock \emph{arXiv preprint arXiv:2002.00240}, 2020.

\bibitem[Paszke et~al.(2017)Paszke, Gross, Chintala, Chanan, Yang, DeVito, Lin,
  Desmaison, Antiga, and Lerer]{paszke2017automatic}
Adam Paszke, Sam Gross, Soumith Chintala, Gregory Chanan, Edward Yang, Zachary
  DeVito, Zeming Lin, Alban Desmaison, Luca Antiga, and Adam Lerer.
\newblock Automatic differentiation in {PyTorch}.
\newblock In \emph{Advances in neural information processing systems}, 2017.

\bibitem[Ronneberger et~al.(2015)Ronneberger, Fischer, and
  Brox]{ronneberger2015u}
Olaf Ronneberger, Philipp Fischer, and Thomas Brox.
\newblock U-net: Convolutional networks for biomedical image segmentation.
\newblock In \emph{International Conference on Medical image computing and
  computer-assisted intervention}, pp.\  234--241. Springer, 2015.

\bibitem[Shamsian et~al.(2021)Shamsian, Navon, Fetaya, and
  Chechik]{shamsian2021personalized}
Aviv Shamsian, Aviv Navon, Ethan Fetaya, and Gal Chechik.
\newblock Personalized federated learning using hypernetworks.
\newblock In \emph{International Conference on Machine Learning}, 2021.

\bibitem[Simonyan \& Zisserman(2015)Simonyan and Zisserman]{simonyan2014very}
Karen Simonyan and Andrew Zisserman.
\newblock Very deep convolutional networks for large-scale image recognition.
\newblock In \emph{{ICLR}}, 2015.

\bibitem[Sitzmann et~al.(2019)Sitzmann, Zollh{\"{o}}fer, and
  Wetzstein]{sitzmann2019scene}
Vincent Sitzmann, Michael Zollh{\"{o}}fer, and Gordon Wetzstein.
\newblock Scene representation networks: Continuous 3d-structure-aware neural
  scene representations.
\newblock In \emph{NeurIPS}, pp.\  1119--1130, 2019.

\bibitem[Stich(2019)]{stich2018local}
Sebastian~U. Stich.
\newblock Local {SGD} converges fast and communicates little.
\newblock In \emph{International Conference on Learning Representations}, 2019.
\newblock URL \url{https://openreview.net/forum?id=S1g2JnRcFX}.

\bibitem[Suarez(2017)]{suarez2017character}
Joseph Suarez.
\newblock Character-level language modeling with recurrent highway
  hypernetworks.
\newblock In \emph{Proceedings of the 31st International Conference on Neural
  Information Processing Systems}, pp.\  3269--3278, 2017.

\bibitem[Targ et~al.(2016)Targ, Almeida, and Lyman]{targ2016resnet}
Sasha Targ, Diogo Almeida, and Kevin Lyman.
\newblock Resnet in resnet: Generalizing residual architectures.
\newblock \emph{CoRR}, abs/1603.08029, 2016.

\bibitem[Wang et~al.(2017)Wang, Peng, Lu, Lu, Bagheri, and
  Summers]{wang2017chestxray}
Xiaosong Wang, Yifan Peng, Le~Lu, Zhiyong Lu, Mohammadhadi Bagheri, and Ronald
  Summers.
\newblock Chestx-ray8: Hospital-scale chest x-ray database and benchmarks on
  weakly-supervised classification and localization of common thorax diseases.
\newblock In \emph{2017 IEEE Conference on Computer Vision and Pattern
  Recognition(CVPR)}, pp.\  3462--3471, 2017.

\bibitem[Xu et~al.(2019)Xu, Hu, Leskovec, and Jegelka]{xu2018powerful}
Keyulu Xu, Weihua Hu, Jure Leskovec, and Stefanie Jegelka.
\newblock How powerful are graph neural networks?
\newblock In \emph{International Conference on Learning Representations}, 2019.
\newblock URL \url{https://openreview.net/forum?id=ryGs6iA5Km}.

\bibitem[Yang et~al.(2019)Yang, Liu, Chen, and Tong]{yang2019federated}
Qiang Yang, Yang Liu, Tianjian Chen, and Yongxin Tong.
\newblock Federated machine learning: Concept and applications.
\newblock \emph{ACM Transactions on Intelligent Systems and Technology (TIST)},
  10\penalty0 (2):\penalty0 1--19, 2019.

\bibitem[Yehudai et~al.(2021)Yehudai, Fetaya, Meirom, Chechik, and
  Maron]{yehudai2021local}
Gilad Yehudai, Ethan Fetaya, Eli Meirom, Gal Chechik, and Haggai Maron.
\newblock From local structures to size generalization in graph neural
  networks, 2021.

\bibitem[Yurochkin et~al.(2019)Yurochkin, Agarwal, Ghosh, Greenewald, Hoang,
  and Khazaeni]{yurochkin2019bayesian}
Mikhail Yurochkin, Mayank Agarwal, Soumya Ghosh, Kristjan Greenewald,
  Trong~Nghia Hoang, and Yasaman Khazaeni.
\newblock Bayesian nonparametric federated learning of neural networks, 2019.

\bibitem[Zhang et~al.(2020)Zhang, Ren, and Urtasun]{zhang2020graph}
Chris Zhang, Mengye Ren, and Raquel Urtasun.
\newblock Graph hypernetworks for neural architecture search, 2020.

\bibitem[Zhang et~al.(2021)Zhang, Sapra, Fidler, Yeung, and
  Alvarez]{zang2021personalized}
Michael Zhang, Karan Sapra, Sanja Fidler, Serena Yeung, and Jose~M. Alvarez.
\newblock Personalized federated learning with first order model optimization.
\newblock In \emph{International Conference on Learning Representations}, 2021.

\bibitem[Zhao et~al.(2018)Zhao, Li, Lai, Suda, Civin, and
  Chandra]{zhao2018federated}
Yue Zhao, Meng Li, Liangzhen Lai, Naveen Suda, Damon Civin, and Vikas Chandra.
\newblock Federated learning with non-iid data.
\newblock \emph{arXiv preprint arXiv:1806.00582}, 2018.

\bibitem[Zhu et~al.(2020)Zhu, Kairouz, McMahan, Sun, and Li]{zhu2020federated}
Wennan Zhu, Peter Kairouz, Brendan McMahan, Haicheng Sun, and Wei Li.
\newblock Federated heavy hitters discovery with differential privacy.
\newblock In \emph{International Conference on Artificial Intelligence and
  Statistics}, pp.\  3837--3847. PMLR, 2020.

\end{thebibliography}
\bibliographystyle{iclr2022_conference}
\clearpage
\appendix
\section{Network 
architectures}\label{supp:arch}

\begin{figure}[h]
    \centering
    \includegraphics[width=0.95\textwidth]{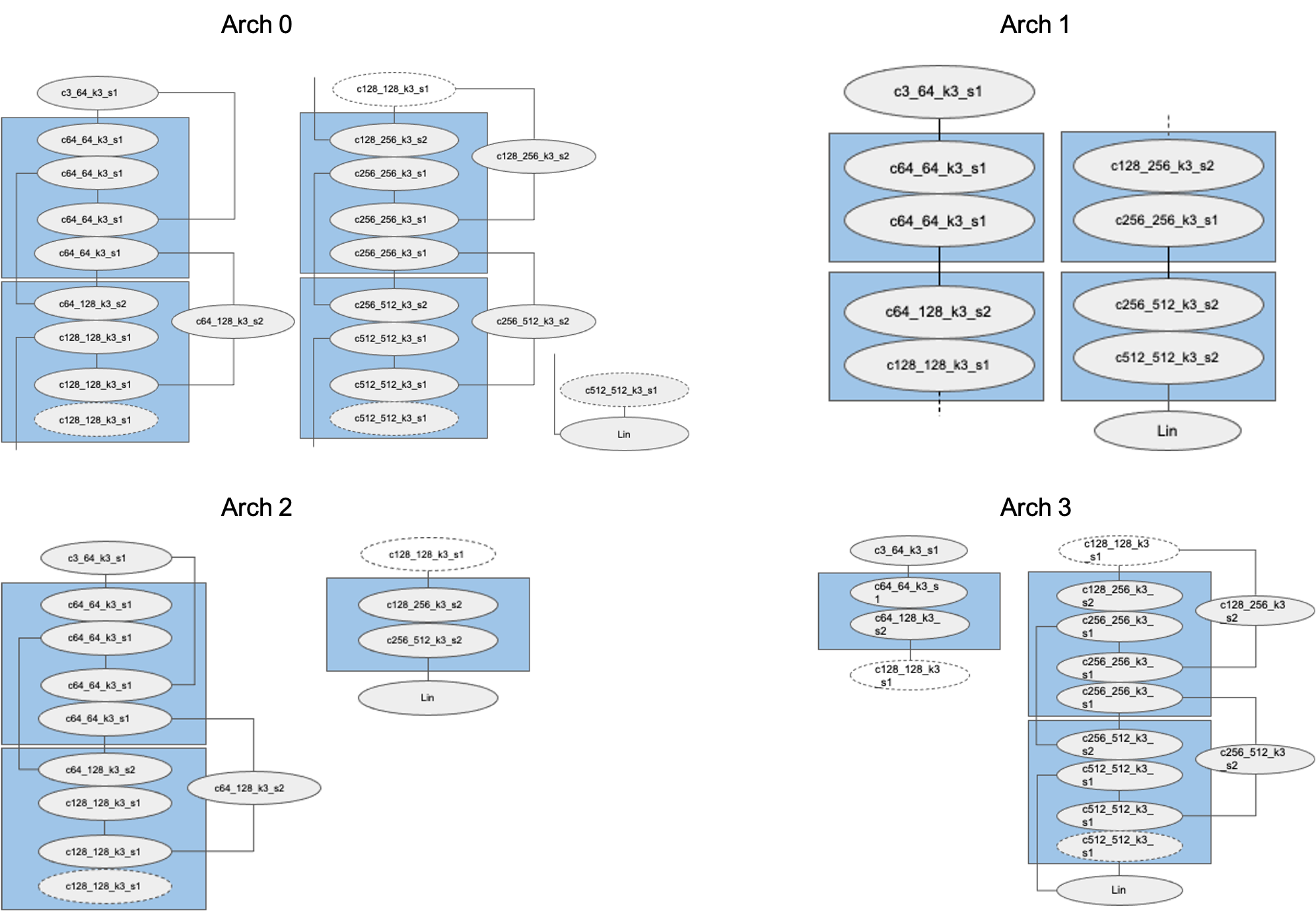}
    \caption{The four client architectures used in CIFAR-10/100 experiments.}
    \label{fig:archs}
\end{figure}

Each experiment in this paper used four different types of architectures split among the different clients \rev{plus an additional small architecture for the stress test}. There are ten different types of nodes (layers) in each architecture.
Figure~\ref{fig:archs} shows the architectures used in the CIFAR-10/100 experiments. The types of convolutional layers are denoted by ``$\text{ c<channels\_in>\_<channels\_ out>\_<kernel\_size>\_<stride>}$''. In the chest x-ray experiment, where the input images are grayscale and of size $224\times 224$ the first convolutional layer type is  ``c1\_64\_k7\_s2'' type. 

CIFAR-10, CIFAR-100, and Chest X-rays, respectively, use different types of linear layers with 10, 100, and 14 output dimensions. 

\section{Non parametric layers}
\rev{
Since our main architectures all use residual connections and same activation type, the graph connectivity suffices to express the architecture. However, in general, different non-parametric layers might be desired. For that, we implemented three versions of ResNet in which several different residual layers were replaced by concatenation, resulting in a hybrid addition-concatenation. 
To that end, we included two new non-parametric types of nodes in our layer type set: “add” and “concat”. Participating in message passing, these nodes produce latent embedding, depending on where they reside in the graph. No additional parameters are required for them. We trained these 3 architectures together with the vanilla ResNet with four and eight clients with the CIFAR10 and CIFAR100 datasets. On CIFAR10, the average results by architecture type are: 88.96, 89, 89, 89.2 for 4 clients and 88.6, 88.1, 88.3, 86.8 for 8 clients. On CIFAR100, the results are: 56, 56.2, 56.5, 48.200 for 4 clients and 50.6, 50.4, 47.7, 46.0 for 8 clients. The results are on par with those shown in Table 1.  In particular, the performance obtained by the vanilla ResNet architecture on CIFAR10/100 with 4 and 8 clients respectively is: 88.96, 88.6, 56 and 50.6, whereas its performance under the \ShortName-distillation baseline is 90, 85.7, 51.1, 44.2. In this example, \ShortName-GHN shows success in training with two commonly used non-parametric layers.}

\section{Hypernetwork initialization}\label{hnet_weight_init_supp}
\begin{figure}[h]
    \centering
    \includegraphics[width=0.95\textwidth]{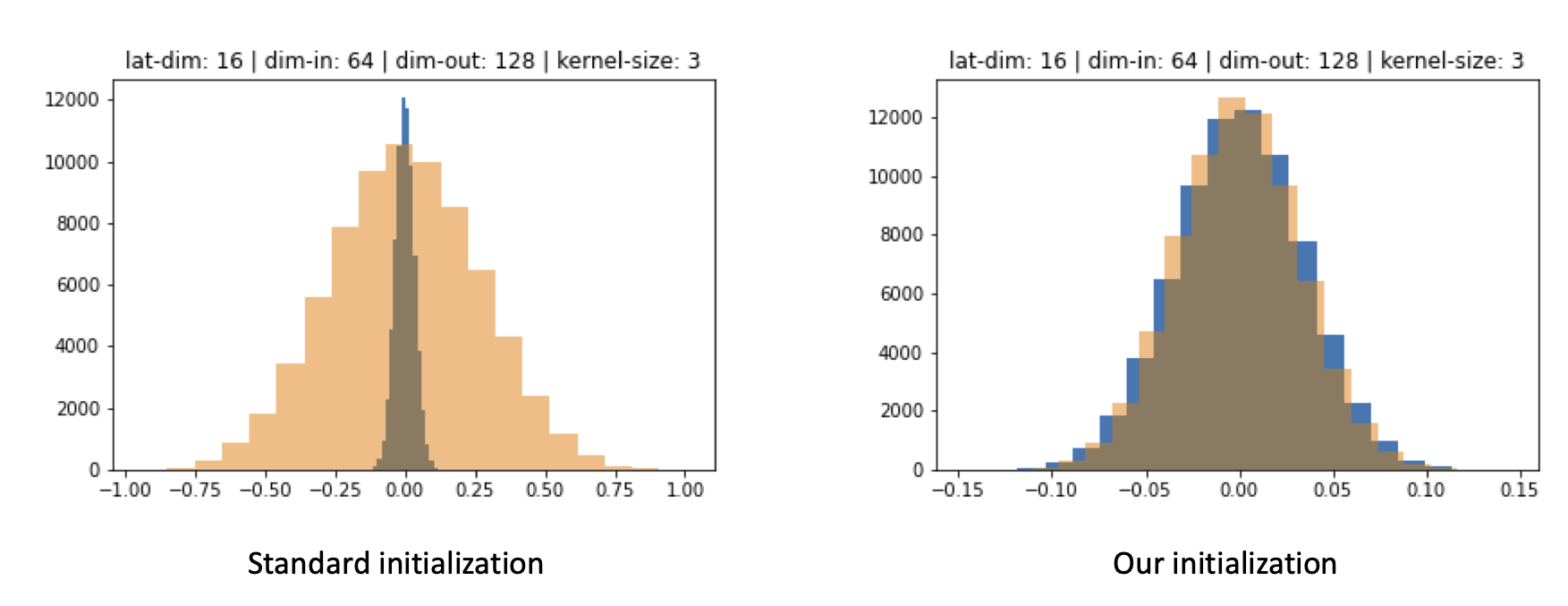}
    \caption{Hypernetwork weight initialization. We compare (blue:) a direct Kaiming weight initialization~\cite{he2015delving} of a convolutional layer with (orange:) the resulting weight initialization by the hypernetwork, without (left) and with (right) our initialization scheme.}
    \label{fig:weight_init}
\end{figure}

As described in Section 4.2 of the main paper, a proper initialization of the hypernetwork weights is instrumental to a successful training of the client networks. In figure~\ref{fig:weight_init} this is shown on an example convolutional layer with dimensions: $3\times 3\times 64\times 128$. In both plots, the blue colored histogram shows the distribution of the desired Kaiming weight initialization~\cite{he2015delving}. When the weights are generated by a hypernetwork, a standard initialization of the hypernetwork would generate the orange histogram shown on the left. We show what that histogram looks like after our initialization scheme on the right. 

\section{hyperarameter search for our \ShortName-GHN}\label{hparamsearch_FLHA}
We ran an extensive hyperparameter search using \cite{wandb} for a 4 architecture setup using a fixed number of 500 epochs , with 3 different GNN types: GraphConv\citep{morris2019weisfeiler}, GatedGraphConv\citep{li2017gated}, and GraphSAGE\citep{hamilton2018inductive}; number of GNN layers $T$ from 1 to 8 with latent dimensions between 16 and 128; hypernetwork $H_l$ bottleneck dimension between 16 and 64; learning rates between 1e-4 and 0.1; SGD (with and without cosine scheduler) and Adam\citep{kingma2014adam} optimizers with weight decay values between 5e-4 and 5e-6. 

\section{Implementation details: Local Distillation}\label{imp_details_local}
Baseline distillation from a teacher model trained via standard (same-architecture) FL is done using a distillation loss~\cite{hinton2015distilling}: $(1- \alpha) \textrm{CE}(y_{pred}, y) + \alpha \textrm{KL}(y_{pred}, y_{teacher}) \times 2T^2$ where CE and KL denote Cross-Entropy and Kullback Leibler, respectively. The softmax in the KL loss is taken with respect to a temperature $T=20.0$ and $\alpha=0.7$. We trained distillation as well as the main FL models for 200 epochs with SGD. The learning rate for training the teacher network (with same-architecture FL) is set to $0.1$ with cosine scheduling. 
When distilling from the teacher network to the student, we use a learning rate to $0.01$. %

\section{Implementation details: pFedHN}\label{imp_details_pfedhn}
We found this architecture to be quite sensitive to hyperparameters, hence unlike our method which uses the same set of parameters in all experiments, here we chose the best performing parameters per setting. Hyperparameters sweep on the following parameters: learning rate, number of hidden layers in shared mlp, latent dimension size, optimizer (adam and sgd), and weight decay.

\section{Unbalanced distribution}\label{subsec:unbalanced}
In collaborative training between different entities, clients' data may be distributed unevenly. In medical data, for example, this may occur when clinics specialize in certain diseases or use different sensors. Thus, in addition to architectural differences, \ShortName can also have data imbalance. Here we study the behavior of \ShortName under such unbalanced data distributions.  
\begin{table}[]
    \centering
    \begin{tabular}{c}
    \includegraphics[width=0.95\textwidth]{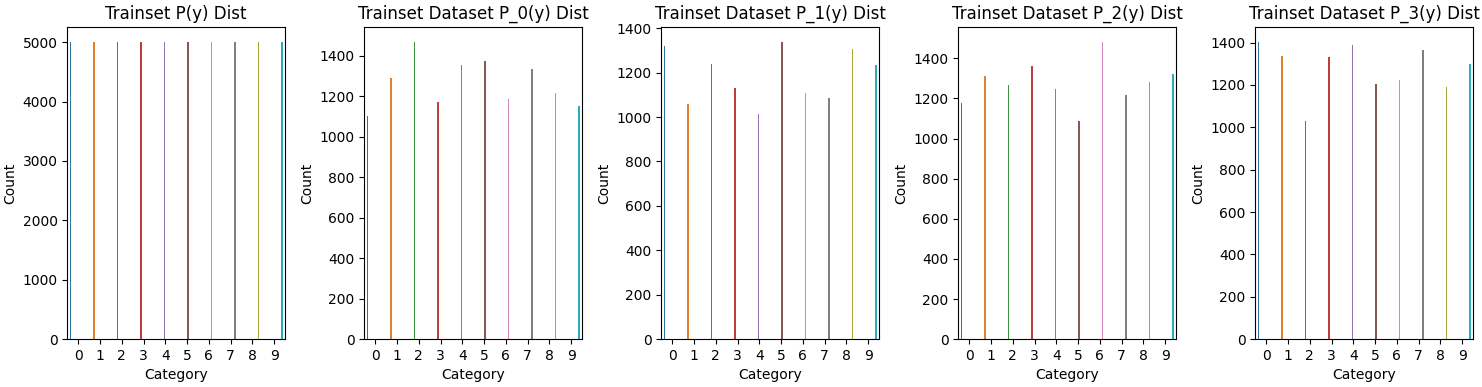} \\
    \midrule
    \includegraphics[width=0.95\textwidth]{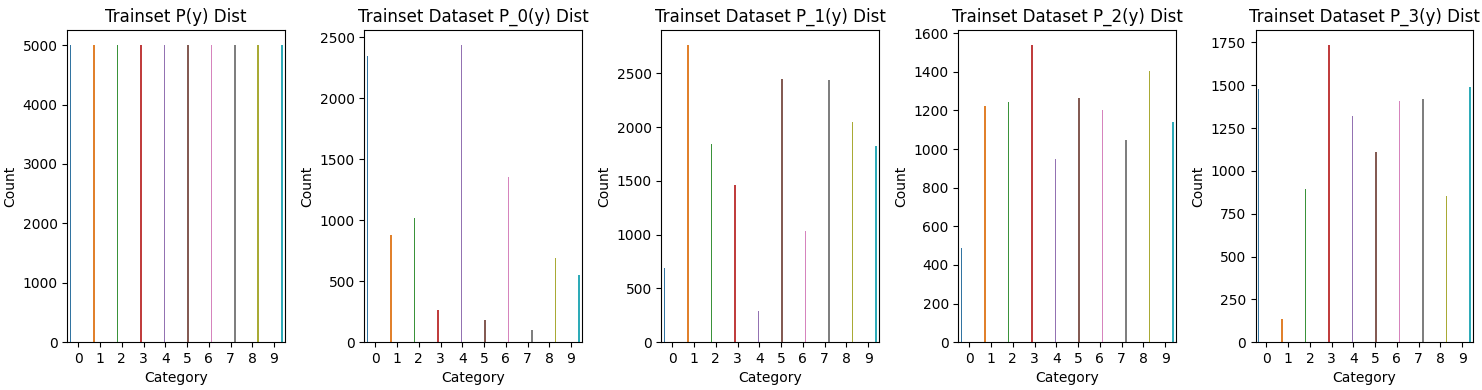} \\
    \midrule
    \includegraphics[width=0.95\textwidth]{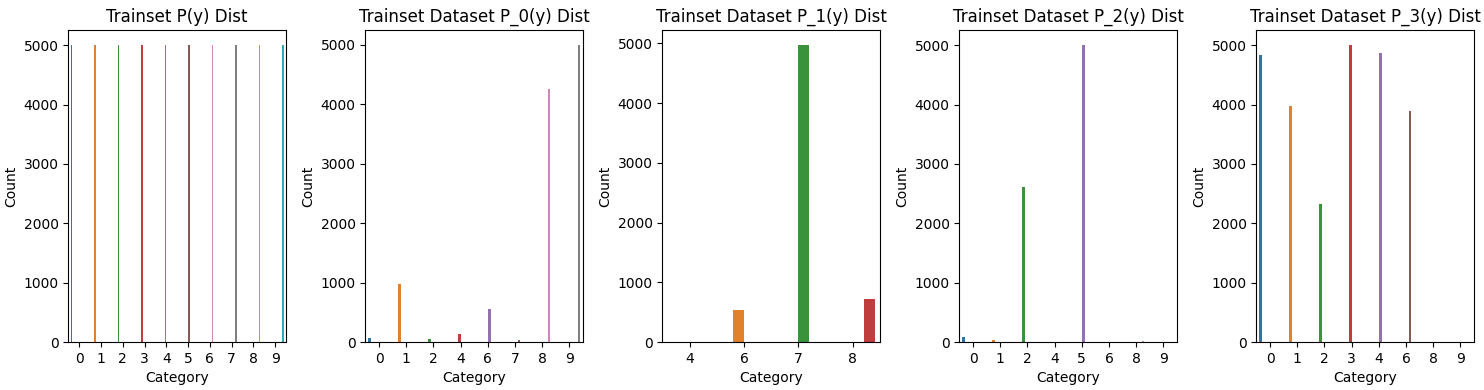}
    \end{tabular}
    \caption{Each row in the table shows the unbalanced class distribution for 4 clients, with the original balanced distribution of the left. Three different $\alpha$ values are shown: (top) $\alpha=100$, (mid) $\alpha=1$, (bottom) $\alpha=0.1$}
    \label{tab:unbalanced_alpha}
\end{table}

We follow \cite{yurochkin2019bayesian,hsu2019measuring,lin2020ensemble} and use the (symmetric) Dirichlet distribution, parameterized by a concentration parameter $\alpha$ to
split the CIFAR-10 training and test sets between the different clients. Table~\ref{tab:unbalanced_alpha} shows the resulting per-client class distributions. As can be seen, the smaller $alpha$ is, the less balanced the distribution is.

Table~\ref{tab:unbalanced} shows the performance on CIFAR-10 under 3 different $\alpha$ values: $0.1, 1, 100$. The smaller $\alpha$ is the more unbalanced the distribution is. Class distribution under the different  $\alpha$ values are shown in Table~\ref{tab:unbalanced_alpha}. 
\begin{table}[htbp]
\footnotesize
  \centering
  \caption{\ShortName with unbalanced distribution. In the table, $\alpha$ corresponds to the level of unbalanced,e.g. $\alpha$=100 (almost uniform),  $\alpha$=0.1 (extremelly unbalanced). unbal. and bal. are short for unbalanced and balanced and correspond to the distribution of the test set with unbalanced being the same distribution of each client's training set.%
  }
    \begin{tabular}{cl|rrrrrrrr}
    \toprule
          &       & \multicolumn{2}{c}{Client 0} & \multicolumn{2}{c}{Client 1} & \multicolumn{2}{c}{Client 2} & \multicolumn{2}{c}{Client 3} \\
    \midrule
    $\alpha$ & method & \multicolumn{1}{c}{unbal.} & \multicolumn{1}{c}{bal.} & \multicolumn{1}{c}{unbal.} & \multicolumn{1}{c}{bal.} & \multicolumn{1}{c}{unbal.} & \multicolumn{1}{c}{bal.} & \multicolumn{1}{c}{unbal.} & \multicolumn{1}{c}{bal.} \\
    \midrule
    \multirow{3}[2]{*}{100} & \ShortName-GHN  & 89.7  & 89.5  & 87.6  & 87.6  & 87.0  & 86.2  & 88.3  & 88.1 \\
          & Local & 81.8  & 82.4  & 75.1  & 74.2  & 80.5  & 80.5  & 81.1  & 80.3 \\
          & Standard FL & 92.9  & 93.7  & 93.2  & 93.7  & 94.1  & 93.7  & 94.5  & 93.7 \\
    \midrule
    \multirow{3}[2]{*}{1} & \ShortName-GHN  & 91.1  & 87.1  & 87.4  & 85.5  & 84.9  & 84.6  & 85.0  & 85.5 \\
          & Local & 87.8  & 75.8  & 87.3  & 83.5  & 84.4  & 84.0  & 84.4  & 83.5 \\
          & Standard FL & 93.8  & 93.0  & 92.9  & 93.0  & 92.6  & 93.0  & 92.9  & 93.0 \\
    \midrule
    \multirow{3}[2]{*}{0.1} & \ShortName-GHN  & 94.6  & 62.2  & 98.3  & 28.7  & 92.3  & 34.0  & 92.4  & 61.0 \\
          & Local & 93.6  & 51.1  & 96.6  & 27.8  & 90.9  & 25.2  & 90.7  & 54.1 \\
          & Standard FL & 69.3  & 53.8  & 58.2  & 53.8  & 42.3  & 53.8  & 49.6  & 53.8 \\
    \bottomrule
    \end{tabular}%
  \label{tab:unbalanced}%
\end{table}%

When the training data is unevenly distributed, performance can either be measured in a similar distributed test set or a balanced test set. In the former case, a client would like high performance on local, biased samples. Suppose a hospital specializes in a specific disease and hopes to improve model performance related to that disease through FL. We refer to that a ``unbalanced'' metric. Alternatively, a client might be interested in balancing its model bias, in which case the performance on the full (``balanced'') test set is of interest. We compare our \ShortName-GHN against a local training. We also include the usual upper bound performance of standard FL with all clients using the same architecture. Table~\ref{tab:unbalanced} shows that \ShortName-GHN outperforms local training in both ``unbalanced'' and ``balanced'' tasks and across all $alpha$ values. Specifically, it can be seen that local training achieves high performance only on test sets of similar distribution, but severely sacrifices performance on sets of balanced distributions. The same architecture FL also shows a trade-off. Despite being the most performant on the balanced test set, it sacrifices accuracy on local distributions. \ShortName-GHN's capability to improve on local training in both tasks can be attributed to the inherent personalization of the network. That is, beyond its ability to adapt to new architectures, \ShortName-GHN can also learn a personalized weight prediction according to the client distributions. This result aligns well with the observation of \cite{shamsian2021personalized}.

\section{Generalization -- additional results}
Table \ref{tab:small_arch}  shows the performance of the 4 architectures used in our experiments, and how they are influenced when each is replaced by a smaller architecture. The average drop in performance of $2.2 \pm 1.4$ keep the performance well above the local-training alternative.

\begin{table}[]
     \vspace{-2mm}
\footnotesize
  \centering
  \caption{Generalization to unseen architectures: leave-one-architecture-out experiment on CIFAR-10. Each row is the accuracy on a held-out architecture while training on the other architectures.}
    \begin{tabular}{lrr}
    \toprule
          & \multicolumn{1}{l}{\textcolor[rgb]{ .114,  .11,  .114}{GHN Init }} & \multicolumn{1}{l}{\textcolor[rgb]{ .114,  .11,  .114}{ From scratch}} \\
    \midrule
    Arch 1 (original) & \textcolor[rgb]{ .114,  .11,  .114}{86.1} & \textcolor[rgb]{ .114,  .11,  .114}{83.9} \\
    Arch 2 (No skip)& \textcolor[rgb]{ .114,  .11,  .114}{84.2} & \textcolor[rgb]{ .114,  .11,  .114}{81.3} \\
    Arch 3 (Skip first)& \textcolor[rgb]{ .114,  .11,  .114}{84.3} & \textcolor[rgb]{ .114,  .11,  .114}{83.7} \\
    Arch 4 (Skip last)& \textcolor[rgb]{ .114,  .11,  .114}{85.6} & \textcolor[rgb]{ .114,  .11,  .114}{83.6} \\
    \bottomrule
    \end{tabular}%
  \label{tab:gen_arch}%
  \vspace{-2mm}

\end{table}

\begin{table}[htbp]
  \centering
  \caption{Training with a much smaller architecture shows an average performance drop by $2.2 \pm 1.4$ pts. However, this is  well above the local training alternative. }
    \begin{tabular}{c|cccc}
    \toprule
    Replaced & Arch 1 & Arch 2 & Arch 3 & \multicolumn{1}{c}{Arch 4} \\
    \midrule
    None  & 90.4  & 89.0  & 87.3  & 88.5 \\
    Arch 4 & 88.2  & 86.9  & 86.4  & \cellcolor[rgb]{ .906,  .902,  .902}80.3 \\
    Arch 3 & 88.8  & 87.1  & \cellcolor[rgb]{ .906,  .902,  .902}79.6 & 88.2 \\
    Arch 2 & 88.6  & \cellcolor[rgb]{ .906,  .902,  .902}81.6 & 85.4  & 86.2 \\
    Arch 1 & \cellcolor[rgb]{ .906,  .902,  .902}77.5 & 83.8  & 85.4  & 83.6 \\
    \bottomrule
    \end{tabular}%
  \label{tab:small_arch}%
\end{table}%

\begin{figure}
    \begin{center}
    \includegraphics[width=0.6\textwidth]{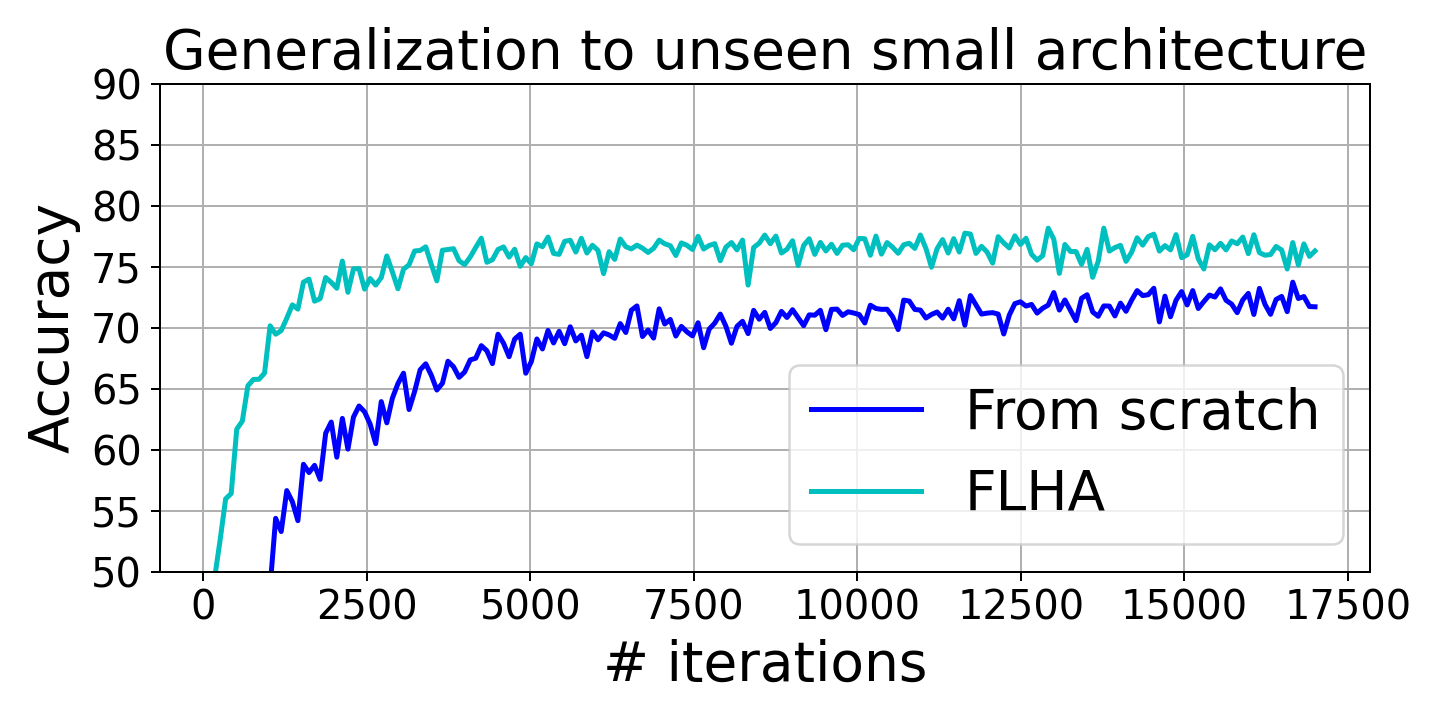}
  \end{center}
  \caption{Generalization to a smaller 4 layer CNN architecture. Our method (light blue) quickly ramps up to high performance and maintains a considerable gap compared to training from scratch until convergence(dark blue).}
  \label{fig:small_arch}

\end{figure}

\end{document}